\documentclass[journal]{IEEEtran}
\usepackage{amsmath,amsfonts}
\usepackage{algorithmic}
\usepackage{algorithm}
\usepackage{array}
\usepackage[caption=false,font=normalsize,labelfont=sf,textfont=sf]{subfig}
\usepackage{textcomp}
\usepackage{stfloats}
\usepackage{url}
\usepackage{verbatim}
\usepackage{graphicx}
\usepackage{cite}
\usepackage{booktabs}
\usepackage{multirow}
\usepackage{longtable}
\usepackage{tablefootnote}
\usepackage{cuted}
\usepackage{tabularx}
\usepackage{makecell}
\usepackage{placeins}
\usepackage[table]{xcolor}

\usepackage{threeparttable}
\usepackage{pifont}
\usepackage{hyperref}
\newcommand{\xmark}{\ding{55}}%
\newcommand{\circleddash}{$\circ$}
\newcommand{\texticon}{\includegraphics[height=0.3cm]{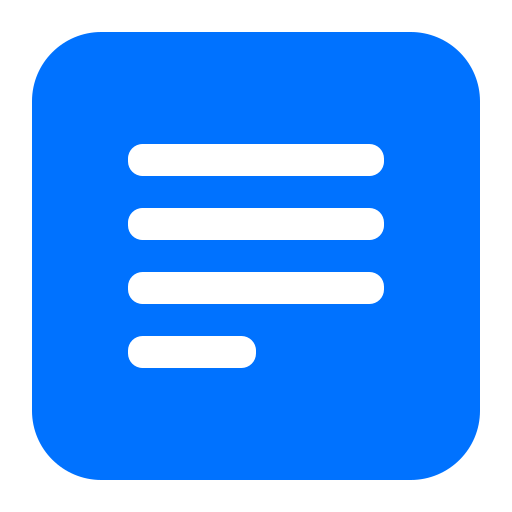}}
\newcommand{\videoicon}{\includegraphics[height=0.3cm]{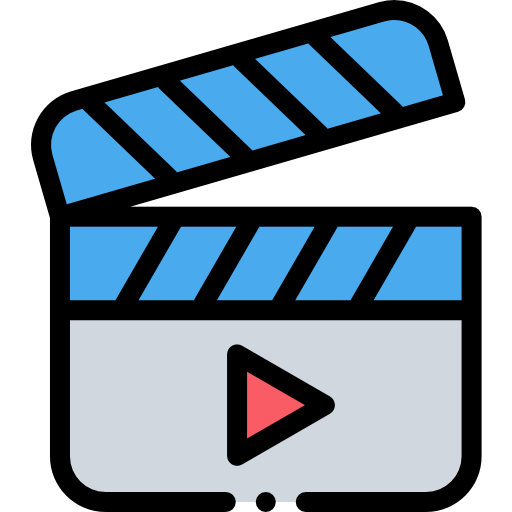}}
\newcommand{\audioicon}{\includegraphics[height=0.3cm]{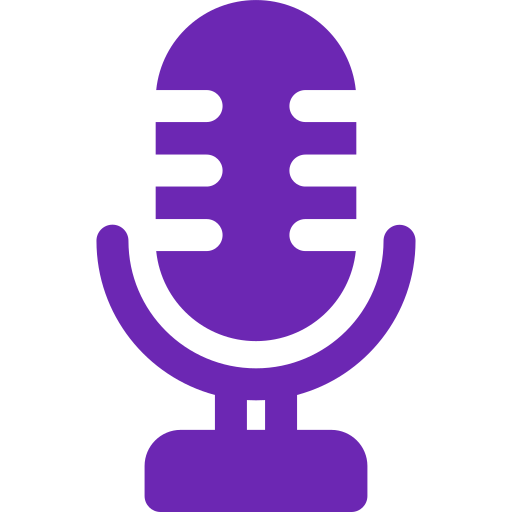}}
\newcommand{\imageicon}{\includegraphics[height=0.3cm]{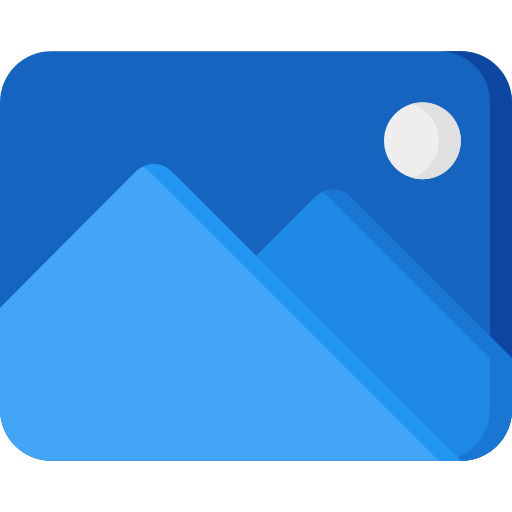}} 
\newcommand{\faceicon}{\includegraphics[height=0.3cm]{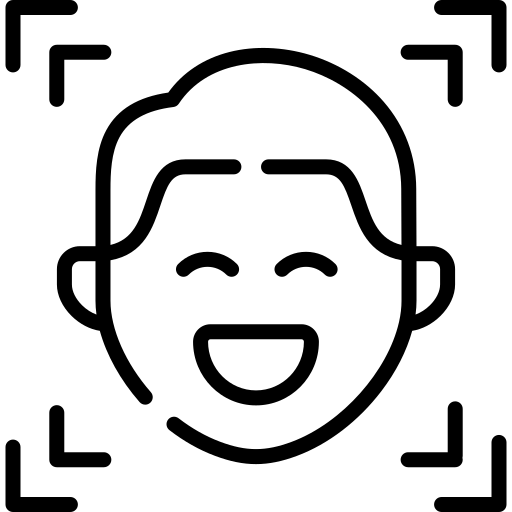}} 
\newcommand{\bodyicon}{\includegraphics[height=0.3cm]{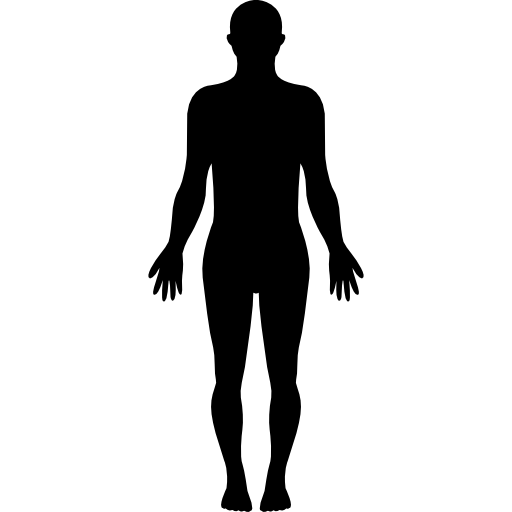}} 
\newcommand{\motionicon}{\includegraphics[height=0.3cm]{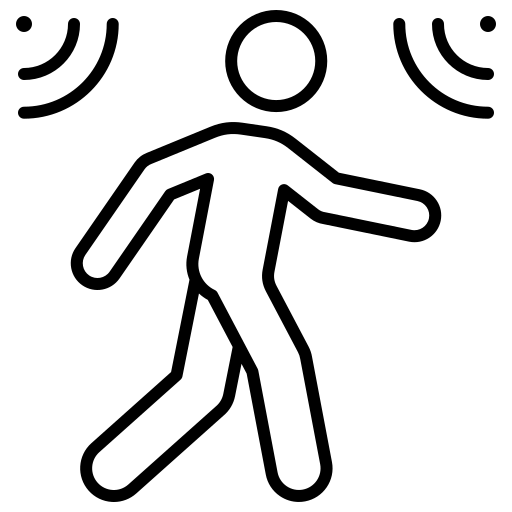}} 

\newcolumntype{C}{>{\centering\arraybackslash}X}

\begin{document}

\title{A Survey of Body and Face Motion: Datasets, Performance Evaluation Metrics and Generative Techniques}

\author{Lownish Rai Sookha,~\IEEEmembership{Student Member,~IEEE}
Nikhil Pakhale,
Mudasir Ganaie and
Abhinav Dhall,~\IEEEmembership{Member,~IEEE}
\thanks{Lownish Rai Sookha, Nikhil Pakhale and Mudasir Ganaie are with the Indian Institute of Technology Ropar, India}
\thanks{Abhinav Dhall is with the Monash University, Australia}%
\thanks{Manuscript received February 18, 2026}}

\markboth{IEEE Transactions on Visualization and Computer Graphics}%
{Shell \MakeLowercase{\textit{et al.}}: A Sample Article Using IEEEtran.cls for IEEE Journals}


\maketitle

\begin{abstract}
Body and face motion play an integral role in communication. They convey crucial information on the participants. Advances in generative modeling and multi-modal learning have enabled motion generation from signals such as speech, conversational context and visual cues. However, generating expressive and coherent face and body dynamics remains challenging due to the complex interplay of verbal/non-verbal cues and individual personality traits. This survey reviews body and face motion generation, covering core concepts, representations techniques, generative approaches, datasets and evaluation metrics. We highlight future directions to enhance the realism, coherence and expressiveness of avatars in dyadic \textbf{(two-person conversational)} settings. \textbf{To the best of our knowledge, this work is the first comprehensive review to cover both body and face motion.} Detailed resources are listed on \href{https://lownish23csz0010.github.io/mogen/}{https://lownish23csz0010.github.io/mogen/}.
\end{abstract}


\section{Introduction}
\label{sec:intro}
\begin{figure*}[!b]
    \centering
    \includegraphics[width=1.0\linewidth]{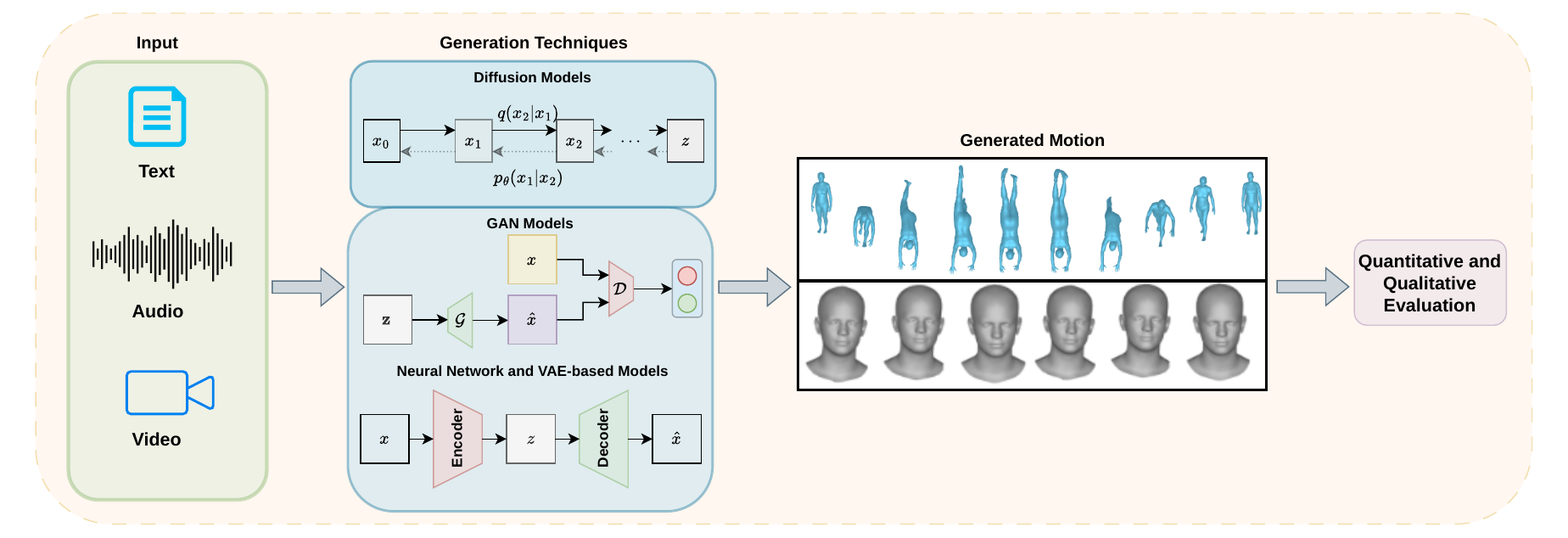}
    \caption{Overview of Generic Motion Generation Pipeline of existing SOTAs. Given the input from the respective modalities, the methods generate desired body or face motion using appropriate representation techniques.}
    \label{fig:pipeline}
\end{figure*}
Human communication and expression extend beyond speech to encompass a rich interplay of verbal and non-verbal behaviors, including facial expressions, gestures, body movements and their coordination with external stimuli such as music and objects. These multi-modal cues convey emotional states, intentions and levels of engagement, forming the basis of natural, synchronous interaction between individuals~\cite{kendon1970movement}. Subtle actions such as head nods, rhythmic movements, or object manipulations reflect attentiveness, affect and context awareness, contributing to the fluidity and meaning of interaction. With the growing advancements in Artificial Intelligence (AI) and Machine Learning (ML), recent research has focused on modeling and generating such complex motion behaviors, ranging from conversational gestures and facial expressions to dance and object-driven actions. Existing surveys typically address either body or face motion generation in isolation, whereas our work provides a unified perspective on both, with a special focus on the evaluation metrics and different generation techniques~\cite{olugbade2022human, liao2024appearance, kammoun2022generative}.

In recent years, the advent of deep learning has led to a surge in generative models, including AutoRegressive (AR) models~\cite{bengio2003neural}, Variational Autoencoders (VAEs)~\cite{DBLP:journals/corr/KingmaW13}, Normalizing Flows~\cite{pmlr-v37-rezende15}, Generative Adversarial Networks (GANs)~\cite{10.1145/3422622} and Denoising Diffusion Probabilistic Models (DDPMs)~\cite{ho2020denoising}. These models have achieved remarkable success across diverse domains such as text, image, video and 3D object generation. Despite these advancements, human motion generation remains a challenging problem due to the complexity of human behavior, the diversity of motion patterns,and the subtle influence of personality, social interaction and contextual cues. Consequently, the field has evolved into a specialized research direction that requires modeling not only realistic motion dynamics but also expressive and socially-aware human behaviors.

\begin{table*}[!ht]
\centering
\caption{Comparison of existing surveys with our survey paper. 
\checkmark~indicates the topic is covered comprehensively, 
\circleddash~indicates partial or limited coverage, and 
\xmark~indicates the topic is not covered.}
\label{tab:survey_comparison}
\renewcommand{\arraystretch}{1.2}
\setlength{\tabcolsep}{5pt}

\begin{tabular}{lccccc}
\toprule
\textbf{Survey Paper} & 
\textbf{Body} & 
\textbf{Face} & 
\textbf{Datasets} & 
\textbf{Evaluation Metrics} & 
\textbf{Unified Coverage} \\
\midrule

Human Motion Generation: A Survey~\cite{zhu2023human} & 
\checkmark & 
\xmark & 
\checkmark & 
\circleddash & 
\xmark \\

Text-driven Motion Generation: Overview, Challenges and Directions~\cite{sahili2025text} & 
\checkmark & 
\xmark & 
\circleddash & 
\circleddash & 
\xmark \\

A Survey on Human Interaction Motion Generation~\cite{sui2026survey} & 
\checkmark & 
\xmark & 
\checkmark & 
\circleddash & 
\xmark \\

Audio-Driven Facial Animation with Deep Learning: A Survey~\cite{jiang2024audio} & 
\xmark & 
\checkmark & 
\checkmark & 
\circleddash & 
\xmark \\

\rowcolor{gray!15}
\textbf{Our Survey} & 
\checkmark & 
\checkmark & 
\checkmark & 
\checkmark & 
\checkmark \\

\bottomrule
\end{tabular}
\end{table*}

To address these challenges, a growing body of research has explored human behavior generation from different perspectives, leading to the emergence of several surveys in related domains. However, existing surveys largely focus on isolated aspects of the problem, leaving a gap for a unified perspective across modalities, interaction settings, and target embodiments. As summarized in Table~\ref{tab:survey_comparison}, Zhu et al.~\cite{zhu2023human} provide a comprehensive overview of body motion generation methods, datasets and evaluation protocols, but their scope is limited to human body motion and does not consider facial behaviors, expressions or conversational dynamics. Similarly, Sahili et al.~\cite{sahili2025text} focus specifically on text-conditioned motion synthesis, restricting their discussion to a single conditioning modality. Sui et al.~\cite{sui2026survey} primarily study physical and multi-person interaction scenarios, but do not adequately address dyadic conversational interaction, where subtle facial expressions, listener responses, gaze and social behaviors play a critical role. Furthermore, Jian et al.~\cite{jiang2024audio} concentrate on audio-driven facial animation, limiting their coverage to facial generation conditioned solely on speech. Collectively, these surveys provide valuable insights into specific sub-domains of human behavior generation, but lack a unified treatment of body and facial motion synthesis across diverse conditioning paradigms and conversational interaction settings.

Motivated by these limitations in existing works, our survey presents a comprehensive review of research contributions spanning from 2016 to 2026, focusing on approaches for human motion synthesis, relevant datasets, and appropriate evaluation metrics. While the survey covers datasets and evaluation protocols introduced between 2016 and 2026, the review of modeling approaches primarily focuses on methods proposed from 2019 onward. This period marks a clear transition in the field, with the widespread adoption of deep generative models, transformer-based architectures, multi-modal learning and, more recently, diffusion models. These developments fundamentally reshaped body and facial motion generation, establishing the methodological foundations that underpin contemporary research. Consequently, the review emphasizes this period while providing historical context where appropriate. In contrast to prior surveys, our work jointly covers both full-body and facial behavior generation, enabling a more holistic understanding of human-centric animation and interaction modeling. We focus on these two aspects because they account for the primary channels through which humans convey actions, emotions and social intent during interactions. Notably, full-body motion generation inherently includes articulated movements of the hands, arms, legs and feet, which are essential for capturing realistic gestures, locomotion and non-verbal communication. Facial behaviors, on the other hand, play a central role in expressing affect, attention and conversational feedback. By jointly studying body and facial motion, the survey aims to provide a unified perspective on human behavior generation across both physical movement and social interaction dynamics.

Moreover, rather than restricting the discussion to a single conditioning modality, our survey encompasses a broader range of modalities including text, audio, visual cues and multi-modal conversational context. We also explicitly incorporate dyadic conversational interaction settings, where facial expressions, listener responses, gaze behavior and social dynamics are central to generating realistic human behaviors. Specifically, our survey covers three primary research domains: \textbf{(1) text-to-motion generation} systems that synthesize 3D human movements from natural language descriptions, \textbf{(2) facial motion generation} models for dyadic conversational settings, and \textbf{(3) multi-modal motion synthesis approaches} that incorporate audio, visual and textual inputs. By unifying body and facial generation, discussing diverse conditioning paradigms, and comprehensively reviewing datasets and evaluation protocols, our survey provides a broader and more integrated perspective on the rapidly evolving landscape of human behavior generation.

The main highlights of this study are as follows:
\begin{enumerate}
    \item We present a comprehensive overview of the datasets used for motion generation, categorized based on their specific application domains.
    
    \item We provide an in-depth summary of the evaluation metrics commonly employed to assess generated motion, organized according to different evaluation criteria.
    
    \item We review recent State-of-the-Art (SOTA) models for both body and facial motion generation, highlighting their key contributions and methodologies.
\end{enumerate}

\section{Preliminaries}
\label{sec:prelims}
\subsection{Human Interaction}

Human interaction in the context of motion generation and conversational AI encompasses the complex dynamics of interpersonal communication, including both verbal and non-verbal elements. The foundation of this research domain builds upon decades of psychological and behavioral studies examining how humans naturally communicate through gesture, facial expression and body language. Dyadic interaction, representing conversation between two participants, forms the core paradigm for many contemporary motion generation systems. Nyatsanga et al.~\cite{nyatsanga2023comprehensive} observe that appropriate facial reactions in conversational settings are inherently non-deterministic, with multiple valid responses possible for any given stimulus. This fundamental insight has shaped the development of stochastic generative models capable of producing diverse, yet contextually appropriate reactions.
The concept of movement coordination in social interaction, originally studied by Kendon~\cite{kendon1970movement}, provides theoretical grounding for understanding how human motions synchronize and complement each other during conversation. This early work continues to influence modern approaches to generate responsive and interactive human avatars. Modern research has expanded this understanding to encompass multi-modal interaction patterns, where facial expressions, head movements and body gestures work in synchrony to convey meaning and maintain engagement. The REACT2023 challenge~\cite{song2023react2023} specifically addresses the generation of multiple appropriate facial reactions, acknowledging that human interaction inherently involves uncertainty and multiple valid response patterns.

\subsection{Body and Face Motion}
Generating body and face motion is a multi-faceted problem. The motion can be the result of a textual description of a sequence of movements~\cite{huang2024como, guo2024momask}, animating a reference image to match the speaker audio~\cite{prajwal2020lip} or the response to certain social cues~\cite{ng2022learning, ng2021body2hands}.
Each of these settings poses distinct challenges. Text-to-motion methods must follow semantic instructions in coherent and physically plausible human movements. Audio-driven animation requires precise temporal alignment between speech features and facial or upper-body articulation, often extending beyond lip motion to capture prosody and expressiveness. Socially conditioned motion generation focuses on modeling human responses to interactive cues, such as gestures or conversational behaviors, requiring an understanding of both dynamics and context.

Together, these modalities illustrate that motion generation encompasses a broad spectrum of tasks with varying inputs and constraints. This diversity shapes the design of current models and provides the backdrop for our study.

\subsection{Representation Techniques}

\begin{figure}[!ht]
    \centering
    \includegraphics[width=1.0\linewidth, height=225pt]{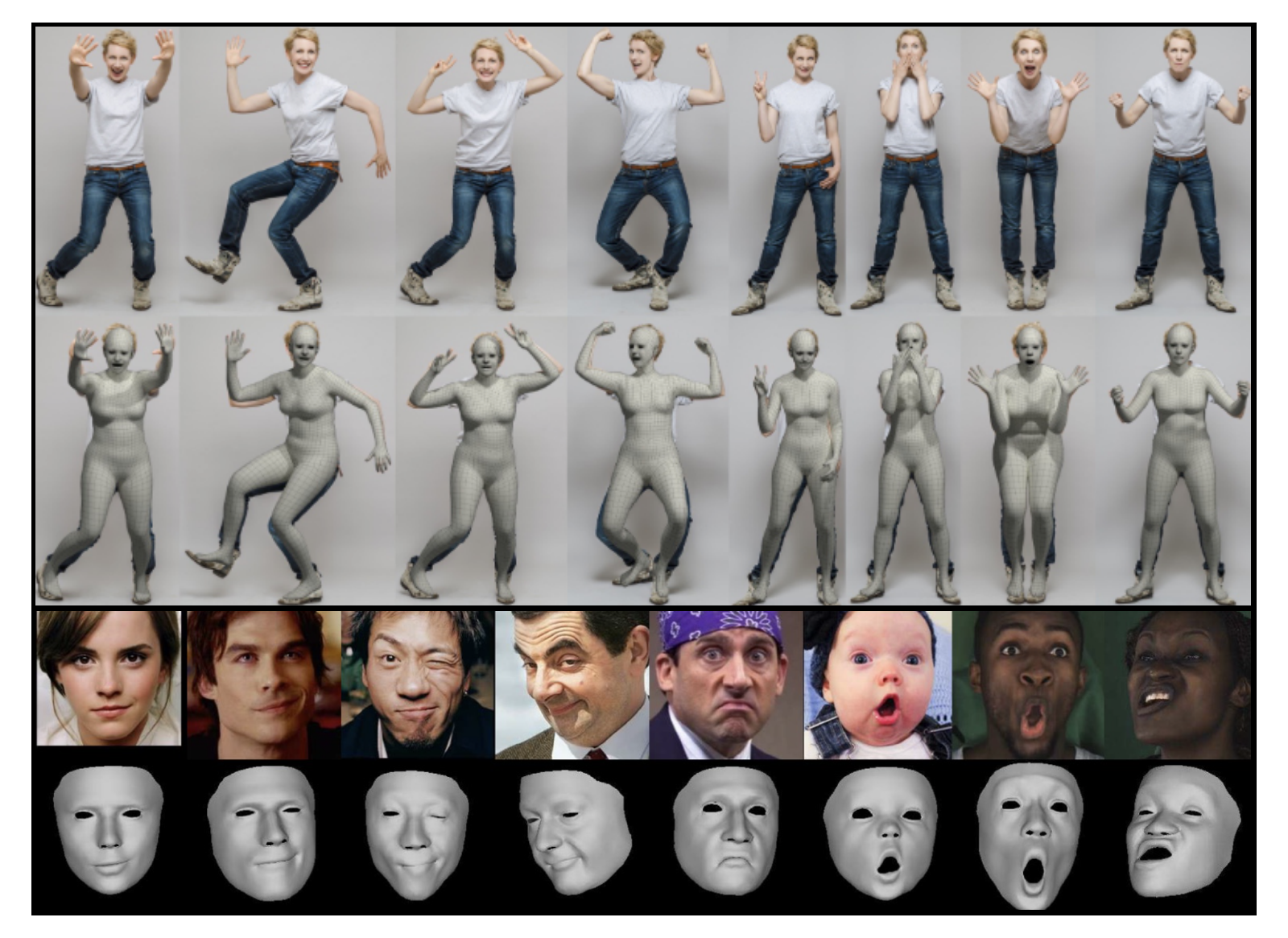}
    \caption{\centering\textbf{(Top)} Given the body image, the body pose and geometry are reconstructed using body representation techniques. Source:~\cite{SMPL-X:2019} \textbf{(Bottom)} Given face images, the face is parameterized and reconstructed using 3DMM frameworks, effectively capturing the pose and expression. Source:~\cite{retsinas20243d}}
    \label{fig:representation}
\end{figure}
Representing human motion is a fundamental challenge, since it requires approaches capable of capturing both the geometric structure and the temporal dynamics of human movement. The methods also need to be computationally efficient for deployment in real life. The frameworks typically decompose human motion into different interpretable components, such as body joints, dependent joint interactions, face pose, expression and texture. While representation in higher dimensions can enable capturing more fine-grained details and more realistic visualizations, it may not always be computationally efficient. Moreover, the level of detail is inherently constrained by the quality and resolution of the input data.

To address these requirements, several parametric modeling frameworks have been developed. The Skinned Multi-Person Linear model family represents a significant advancement in human body representation~\cite{SMPL:2015}. The Master Motor Map (MMM) framework~\cite{mandery2016unifying, terlemez2014master, azad2007toward} provides a unified representation for human motion, while 3D Morphable Models (3DMM)~\cite{3dmm} lay the foundation for parametric modeling later extended to full-body representations. The SMPL-X model~\cite{pavlakos2019expressive} provides a unified representation for body, face and hands, enabling complete motion synthesis of whole body. On the other hand, recent developments in facial representation include DECA~\cite{feng2021learning} for detailed facial expression capture and EMOCA~\cite{danvevcek2022emoca} for emotion-driven face animation. SMIRK~\cite{retsinas20243d} introduces analysis-by-synthesis approaches for 3D facial expression modeling, demonstrating the evolution towards more sophisticated representation techniques. Figure~\ref{fig:representation} shows the body and face reconstruction of existing frameworks.

While these models establish the structural foundation of the human body and face reconstruction, motion representation introduces an additional dimension focused on temporal dynamics. A wide range of strategies has been explored to encode these dynamics. T2M-GPT~\cite{zhang2023generating} introduces vector quantization for discrete body motion representation, while MotionGPT~\cite{jiang2023motiongpt} treats body motion as a foreign language, tokenizing movement patterns for language model processing. Similarly, L2L~\cite{ng2022learning} discretizes face motion using a learnable codebook.

\begin{figure}[!ht]
    \centering
    \includegraphics[width=0.95\linewidth, height=170pt]{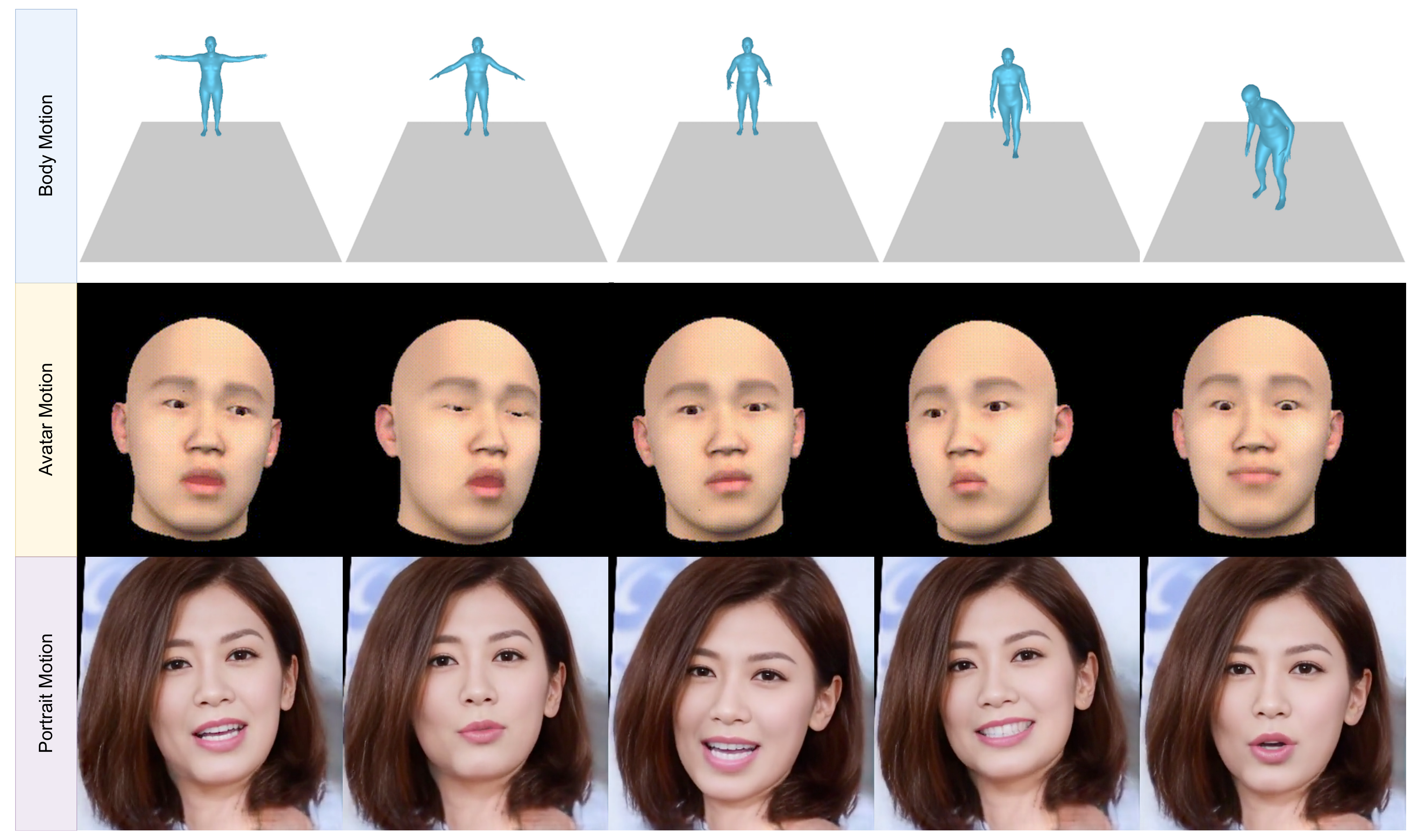}
    \caption{Visualization of existing motion generation frameworks. For better view, kindly visit the \href{https://lownish23csz0010.github.io/mogen/}{Project page}}
    \label{fig:video_snaps}
\end{figure}

Beyond the choice of discrete tokenization, temporal encoding itself presents additional challenges, with researchers exploring various encoding strategies for capturing motion dynamics. Some approaches utilize absolute joint positions over time, while others focus on relative motion patterns or velocity-based representations. The choice of representation has a significant impact on both the quality of generated motions and the computational requirements of the synthesis process. Some visualizations of motion generation SOTAs, in the area of body motion, avatar motion and face portrait motion, are shown in Figure~\ref{fig:video_snaps}.

\section{Datasets}
\begin{table*}[!htbp]
\centering
\begin{threeparttable}[b]

\caption{Datasets for Human Motion Generation\tnote{\ding{61}}}
\label{tab:dataset_tab}

\begin{tabular}{>{\centering\arraybackslash}p{2.5cm}lcccc}
\toprule
\textbf{Category} &
\textbf{Dataset} &
\textbf{Year} &
\textbf{Modality} &
\textbf{Region} &
\textbf{Representation} \\
\midrule

\multirow{7}{*}{\parbox{2.5cm}{\centering\textbf{Text-Conditioned}}} &
MotionFix~\cite{athanasiou2024motionfix} & 2024 & \texticon & \bodyicon & Skeleton \\
& Motion-X~\cite{lin2023motion} & 2023 & \texticon $+$ \videoicon & \bodyicon & Skeleton \\
& CelebV-Text~\cite{yu2023celebv} & 2023 & \texticon $+$ \audioicon $+$ \videoicon & \faceicon & 2D Video \\
& HumanML3D~\cite{guo2022generating} & 2022 & \texticon & \bodyicon & Skeleton \\
& HUMANISE~\cite{wang2022humanise} & 2022 & \texticon & \bodyicon & Skeleton \\
& BABEL~\cite{punnakkal2021babel} & 2021 & \texticon & \bodyicon & Skeleton \\
& KIT-ML~\cite{plappert2016kit} & 2016 & \texticon & \bodyicon & Skeleton \\
\midrule

\multirow{5}{*}{\parbox{2.5cm}{\centering\textbf{Audio-Conditioned}}} &
MOSA~\cite{huang2024mosa} & 2024 & \audioicon $+$ \videoicon & \bodyicon & Skeleton \\
& AIOZ-GDANCE~\cite{le2023music} & 2023 & \audioicon $+$ \videoicon & \bodyicon & 2D Video \\
& PhantomDance~\cite{li2022danceformer} & 2022 & \audioicon $+$ \videoicon & \bodyicon & 2D Video \\
& AIST++~\cite{li2021ai} & 2021 & \audioicon $+$ \videoicon & \bodyicon & Skeleton \\
& AniDance~\cite{tang2018anidance} & 2018 & \audioicon $+$ \videoicon & \bodyicon & 2D Video \\
\midrule

\multirow{19}{*}{\parbox{2.5cm}{\centering\textbf{Speech-Conditioned}}} &
CHDTF~\cite{liu2025disenttalk} & 2025 & \audioicon $+$ \videoicon & \faceicon & 2D Video \\
& Hallo3~\cite{cui2024hallo3} & 2024 & \audioicon $+$ \videoicon & \bodyicon & 2D Video \\
& MultiTalk~\cite{sungbin24_interspeech} & 2024 & \texticon $+$ \audioicon $+$ \videoicon & \bodyicon & 2D Video \\
& ZEGGS~\cite{ghorbani2023zeroeggs} & 2023 & \texticon $+$ \audioicon $+$ \videoicon & \bodyicon $+$ \faceicon & 2D Video \\
& BEAT~\cite{liu2022beat} & 2022 & \texticon $+$ \audioicon $+$ \videoicon & \bodyicon $+$ \faceicon & 2D Video \\
& CelebV-HQ~\cite{zhu2022celebvhq} & 2022 & \texticon $+$ \audioicon $+$ \videoicon & \faceicon & 2D Video \\
& VFHQ~\cite{xie2022vfhq} & 2022 & \audioicon $+$ \videoicon & \faceicon & 2D Video \\
& ViCo~\cite{zhou2022responsive} & 2022 & \audioicon $+$ \videoicon & \faceicon & 3DMM \\
& TalkHead-1KH~\cite{wang2021one} & 2021 & \audioicon $+$ \videoicon & \faceicon & 2D Video \\
& MEAD~\cite{wang2020mead} & 2020 & \audioicon $+$ \videoicon & \faceicon & 2D Video \\
& PATS~\cite{ahuja2020no,ahuja2020style} & 2020 & \texticon $+$ \audioicon $+$ \videoicon & \bodyicon $+$ \faceicon & 2D Video \\
& Trinity~\cite{ferstl2018investigating} & 2020 & \texticon $+$ \audioicon $+$ \videoicon & \bodyicon $+$ \faceicon & 2D Video \\
& Speech2Gesture~\cite{ginosar2019learning} & 2019 & \texticon $+$ \audioicon $+$ \videoicon & \bodyicon $+$ \faceicon & 2D Video \\
& CelebV~\cite{wu2018reenactgan} & 2018 & \audioicon $+$ \videoicon & \faceicon & 2D Video \\
& VoxCeleb2~\cite{chung2018voxceleb2} & 2018 & \audioicon $+$ \videoicon & \faceicon & 2D Video \\
& VoxCeleb1~\cite{nagrani2017voxceleb} & 2017 & \audioicon $+$ \videoicon & \faceicon & 2D Video \\
& LRS~\cite{son2017lip} & 2017 & \audioicon $+$ \videoicon & \faceicon & 2D Video \\
& MV-LRS~\cite{chung2017lip} & 2017 & \audioicon $+$ \videoicon & \faceicon & 2D Video \\
& LRW~\cite{chung2016lip} & 2016 & \audioicon $+$ \videoicon & \faceicon & 2D Video \\
\midrule

\multirow{10}{*}{\parbox{2.5cm}{\centering\textbf{Scene-Conditioned}}} &
Habitat~\cite{puig2023habitat3, szot2021habitat, habitat19iccv} & 2023 & \videoicon & \bodyicon & Skeleton \\
& Circle~\cite{araujo2023circle} & 2023 & \audioicon $+$ \videoicon & \bodyicon & Skeleton \\
& HUMANISE~\cite{wang2022humanise} & 2022 & \texticon & \bodyicon & Skeleton \\
& COUCH~\cite{zhang2022couch} & 2022 & \videoicon & \bodyicon & Skeleton \\
& SAMP~\cite{hassan2021stochastic} & 2021 & \videoicon & \bodyicon & Skeleton \\
& GTA-IM~\cite{cao2020long} & 2020 & \videoicon & \bodyicon $+$ \faceicon & 2D Video \\
& PROX~\cite{hassan2019resolving} & 2019 & \videoicon & \bodyicon & Mesh \\
& JTA~\cite{fabbri2018learning} & 2018 & \videoicon & \bodyicon & 2D Video \\
& PiGraph~\cite{savva2016pigraphs} & 2016 & \texticon $+$ \videoicon & \bodyicon & Skeleton \\
\midrule

\multirow{9}{*}{\parbox{2.5cm}{\centering\textbf{Action-Conditioned}}} &
Motion-X~\cite{lin2023motion} & 2023 & \texticon $+$ \videoicon & \bodyicon & Skeleton \\
& BABEL~\cite{punnakkal2021babel} & 2021 & \texticon & \bodyicon & Skeleton \\
& EMOGAIT~\cite{sheng2021multi} & 2021 & \texticon & \bodyicon & Skeleton \\
& HuMMan~\cite{cai2022humman} & 2021 & \texticon $+$ \videoicon & \bodyicon & Skeleton \\
& HumanAct12~\cite{guo2020action2motion} & 2020 & \texticon & \bodyicon & Skeleton \\
& NTU-RGB+D~\cite{shahroudy2016ntu, liu2019ntu} & 2016 & \texticon $+$ \videoicon & \bodyicon & 2D Video \\
& Penn Action~\cite{zhang2013actemes} & 2013 & \texticon $+$ \videoicon & \bodyicon & 2D Video \\
& UCF101~\cite{soomro2012ucf101} & 2012 & \texticon $+$ \videoicon & \bodyicon & 2D Video \\
\midrule

\multirow{7}{*}{\parbox{2.5cm}{\centering\textbf{General \\Motion Capture}}} &
BioCV~\cite{evans2024biocv} & 2024 & \videoicon & \bodyicon & 2D Video \\
& Motion-X~\cite{lin2023motion} & 2023 & \texticon $+$ \videoicon & \bodyicon & Skeleton \\
& HuMMan~\cite{cai2022humman} & 2021 & \texticon $+$ \videoicon & \bodyicon & Skeleton \\
& MoVi~\cite{ghorbani2020movi} & 2021 & \videoicon & \bodyicon & Skeleton \\
& AMASS~\cite{mahmood2019amass} & 2019 & \texticon & \bodyicon & Mesh \\
& Human3.6M~\cite{h36m_pami} & 2014 & \texticon $+$ \videoicon & \bodyicon & Skeleton \\

\bottomrule

\end{tabular}
\vspace{3pt}

\begin{tablenotes}
\footnotesize
\item[\ding{61}] Some datasets may appear in multiple categories due to their later applications in specialized domains.
\item \texticon: Text; \audioicon: Audio; \videoicon: Video (Excludes audio); \bodyicon: Body; \faceicon: Face
\end{tablenotes}

\end{threeparttable}
\end{table*}

\begin{table*}[!htbp]
\centering
\begin{threeparttable}[b]

\caption{Datasets for Human Motion Generation\tnote{\ding{61}}\textit{(Continued)}}

\begin{tabular}{>{\centering\arraybackslash}p{2.5cm}lcccc}
\toprule
\textbf{Category} &
\textbf{Dataset} &
\textbf{Year} &
\textbf{Modality} &
\textbf{Region} &
\textbf{Representation} \\
\midrule

\multirow{13}{*}{\parbox{2.5cm}{\centering\textbf{Interaction}}} &
MMF2F~\cite{lin-etal-2025-predicting} & 2025 & \texticon $+$ \audioicon $+$ \videoicon & \faceicon & 2D Video \\
& DyConv ~\cite{zhu2025infp} & 2025 & \audioicon $+$ \videoicon & \faceicon & 2D Video \\
& CCDb+ ~\cite{aubrey2013cardiff, VandeventerARM15, deng2025ccdb+} & 2025 & \audioicon $+$ \videoicon & \faceicon & 2D Video \\
& NoXi-J~\cite{funk2024multilingual} & 2024 & \audioicon $+$ \videoicon & \bodyicon & 2D Video \\
& InterHuman~\cite{liang2024intergen} & 2024 & \texticon $+$ \videoicon & \bodyicon & Skeleton \\
& Audio2Photoreal~\cite{ng2024audio2photoreal} & 2024 & \audioicon $+$ \videoicon & \bodyicon $+$ \faceicon & Skeleton \& Mesh \\
& RealTalk~\cite{geng2023affective} & 2023 & \texticon $+$ \audioicon $+$ \videoicon & \faceicon & 3DMM \\
& L2L~\cite{ng2022learning, ng2023can} & 2023 & \texticon $+$ \audioicon $+$ \videoicon & \faceicon & 3DMM \\
& UDIVA~\cite{palmero2021context} & 2021 & \audioicon $+$ \videoicon & \faceicon & 2D Video \\
& Body2Hands~\cite{ng2021body2hands} & 2021 & \videoicon $+$ \audioicon & Skeleton \\
& GRAB~\cite{taheri2020grab} & 2020 & \videoicon & \bodyicon & Mesh \\
& Talking With Hands 16.2M~\cite{lee2019talking} & 2019 & \videoicon $+$ \audioicon & Skeleton \\
& NoXi~\cite{cafaro17_icmi} & 2017 & \audioicon $+$ \videoicon & \bodyicon & 2D Video \\
& RECOLA~\cite{ringeval2013introducing} & 2013 & \audioicon $+$ \videoicon & \faceicon & 2D Video \\
\bottomrule

\end{tabular}
\vspace{3pt}

\begin{tablenotes}
\footnotesize
\item[\ding{61}] Some datasets may appear in multiple categories due to their later applications in specialized domains.
\item \texticon: Text; \audioicon: Audio; \videoicon: Video (Excludes audio); \bodyicon: Body; \faceicon: Face
\end{tablenotes}
\end{threeparttable}
\end{table*}

In this section, we present the datasets commonly used for human motion generation. These datasets are curated for diverse tasks, each emphasizing specific conditioning modalities. We categorize them based on the type of conditioning information they provide as presented in Table~\ref{tab:dataset_tab}.

\subsection{Text-Conditioned Motion Datasets}
Generating motion from text enables fine-grained control over a sequence of actions and is primarily used to synthesize bodily motion. One of the early contributions in this direction is the KIT-ML dataset~\cite{plappert2016kit}, which offers a large-scale marker-based motion capture collection using the MMM framework and establishing a connection between human motion and natural language descriptions. Following this, BABEL~\cite{punnakkal2021babel} provides richer and more fine-grained action annotations for motion clips, significantly improving semantic understanding. One of the latest contribution is HumanML3D~\cite{guo2022generating} which focuses on everyday human actions, further enriching the scope of text-to-motion research.

While most motion datasets emphasize on full-body movement, complementary facial expressions are often overlooked. To address this gap, Motion-X~\cite{lin2023motion} integrates both body and face motion, while MotionFix~\cite{athanasiou2024motionfix} introduces flexible motion editing and HUMANISE~\cite{wang2022humanise} explores scene-aware and context-adaptive actions. In contrast, CelebV-Text~\cite{yu2023celebv} focuses primarily on facial expression and motion, offering rich textual descriptions that capture emotions, appearance, actions and even lighting details.

\subsection{Audio-Conditioned Motion Datasets}
Audio-conditioned motion generation refers to motion synthesis using audio signals, most commonly music. This line of work focuses primarily on producing expressive and synchronized dance movements. Several datasets support this direction, including AniDance~\cite{tang2018anidance}, AIST++~\cite{li2021ai}, PhantomDance~\cite{li2022danceformer} and AIOZ-GDANCE~\cite{le2023music}, each providing rich motion-capture data paired with diverse music tracks, enabling the creation of virtual dancing avatars. While these datasets focus on general dance performance, the MOSA dataset~\cite{huang2024mosa} introduces a unique perspective by modeling the fine-grained motions of musicians as they play various instruments. Instead of broad dance expression, MOSA captures subtle performance cues, such as posture, hand articulation and instrument-specific gestures, highlighting the wide range of motion grounded in audio beyond dance choreography.

\subsection{Speech-Conditioned Motion Datasets}
Speech-driven face generation is a core component of lip-sync tasks. Early contributions in this area leverage datasets such as LRW~\cite{chung2016lip}, MV-LRS~\cite{chung2017lip} and LRS~\cite{son2017lip}, which consist of thousands of English-spoken sentences extracted from BBC television broadcasts. Originally designed for visual speech recognition, mapping lip movements to spoken words, these datasets have since been widely adopted for lip syncing.

To improve diversity across culture, ethnicity and gender, newer datasets such as VoxCeleb1~\cite{nagrani2017voxceleb}, VoxCeleb2~\cite{chung2018voxceleb2}, CelebV~\cite{wu2018reenactgan}, VFHQ~\cite{xie2022vfhq} and CelebV-HQ~\cite{zhu2022celebvhq} are introduced. These datasets include a wide range of celebrity speech videos and are originally developed for tasks such as speaker verification, emotion recognition and face-voice transfer. However, they have since been extended to support face generation and talking-head synthesis. More recent efforts have emphasized higher video quality and longer durations to support high-fidelity face generation.

TalkingHead-1KH~\cite{wang2021one} is among the first datasets to explicitly target talking-head generation using 3DMM coefficients and a reference image, though it is limited to speaking faces. To broaden the scope,  ViCo~\cite{zhou2022responsive} extends to listener motion. Despite its relatively small scale, ViCo sets a benchmark for modeling both speaking and listening behaviors using 3DMM representations. In parallel, MEAD~\cite{wang2020mead} addresses the growing need for emotional expressiveness in talking faces, filling a critical gap in earlier datasets. Moving in this direction, Hallo3~\cite{cui2024hallo3} further enhances realism by supporting fine-grained face animation editing, enabling more dynamic and subtle facial behavior.

Beyond speech-driven face motion, body movement has also been explored. The Trinity dataset~\cite{ferstl2018investigating} focuses on generating realistic speaker body motion during interaction, but is constrained by a single actor and limited diversity. BEAT~\cite{liu2022beat} addresses this with more diverse recordings, emphasizing alignment with both speech semantics and emotion, while also incorporating facial expressions. Speech2Gesture~\cite{ginosar2019learning} and PATS~\cite{ahuja2020no, ahuja2020style} further model full-body gestures exhibited during communication. Most recently, ZEGGS~\cite{ghorbani2023zeroeggs} advances this direction by capturing comprehensive whole-body dynamics, extending beyond arm gestures to more natural movements.

Despite these advancements, most existing datasets are predominantly English-centric, limiting their cross-lingual applicability. MultiTalk~\cite{sungbin24_interspeech} addresses this by introducing a multilingual dataset designed to support the development of language-agnostic models. CHDTF~\cite{liu2025disenttalk}, while primarily focused on Mandarin, also includes English samples for evaluating cross-lingual generalizability.

\subsection{Scene-Conditioned Motion Datasets}

Scene-aware motion generation is an emerging area that focuses on synthesizing human motion in a way that accounts for the surrounding environment and objects. One of the early works in this field is PiGraph~\cite{savva2016pigraphs}, which introduces a probabilistic model that captures the relationship between human poses and object geometries using RGB-D sensors in indoor scenes. Building upon this, the PROX dataset~\cite{hassan2019resolving} enables motion generation that aligns more consistently with 3D environments derived from monocular images, using body poses estimated through SMPL-X~\cite{SMPL:2015} and SMPLify-X~\cite{SMPL-X:2019}. Further incorporating body-scene interaction are the SAMP~\cite{hassan2021stochastic} and Circle~\cite{araujo2023circle} datasets that provide detailed annotations that facilitate learning of context-aware behaviors. The Habitat datasets~\cite{puig2023habitat3, szot2021habitat, habitat19iccv} are designed for high-fidelity simulation of embodied agents, supporting studies in human-robot collaboration within realistic indoor environments. For outdoor motion, the Joint Track Auto (JTA) dataset~\cite{fabbri2018learning}, collected from the photorealistic video game \textit{Grand Theft Auto V}, offers large-scale data for analyzing human movement in complex urban scenarios.

\subsection{Action-Conditioned Motion Datasets}

Action-conditioned motion datasets can be considered a subset of text-conditioned datasets, with the primary difference being the type of textual input. Instead of free-form language, they depend on pre-defined motion labels. UCF101~\cite{soomro2012ucf101} and Penn Action~\cite{zhang2013actemes}, originally developed for action classification, can also be utilized to learn and generate body motion dynamics from short action descriptions. NTU-RGB+D~\cite{shahroudy2016ntu, liu2019ntu} expands this scope by including 82 daily activities, 12 medical-related motions, and 26 two-person interactions, categories largely absent in earlier datasets. It provides rich modalities including RGB videos, depth data and skeleton information. A more refined version of this dataset, HumanAct12~\cite{guo2020action2motion}, enables more fine-grained motion generation.

The HuMMan~\cite{cai2022humman} dataset offers comprehensive multi-modal annotations, including images, point clouds, keypoints, SMPL parameters and textured meshes. With 500 fundamental human actions, it supports tasks such as action recognition, pose estimation, parametric human recovery and textured mesh reconstruction. Complementing this, EMOGAIT~\cite{sheng2021multi} focuses on the interplay between body motion and affect, capturing gait dynamics associated with different emotional states.

\subsection{General Motion Capture Datasets}

Some datasets are designed to serve multiple purposes in human motion research, offering diverse modalities and rich annotations. Motion-X and HuMMan, for instance, provide multi-modal information that supports a wide range of tasks, including pose estimation, action recognition, and motion generation. Similarly, Human3.6M\cite{h36m_pami} is a large-scale benchmark dataset comprising approximately 3.6 million 3D human poses captured in a controlled environment. It features several actors performing a variety of actions from multiple viewpoints, making it especially useful for learning intra-action variations and modeling consistent pose sequences across time. AMASS~\cite{mahmood2019amass} is another widely used large-scale motion capture dataset that unifies numerous MoCap datasets into a common SMPL representation, enabling standardized learning of diverse human motions across subjects, actions and recording conditions. Its broad coverage and consistent parameterization make it particularly valuable for motion generation, motion retargeting and representation learning tasks. Another versatile dataset is MoVi~\cite{ghorbani2020movi}, which captures full-body human motion using synchronized multi-modal recordings such as RGB videos, depth maps, inertial measurements and 3D poses. MoVi is particularly suitable for learning cross-modal representations of motion and analyzing human movement from multiple perspectives. BioCV~\cite{evans2024biocv} is a specialized dataset designed to analyze individual movement patterns with the goal of identifying injury risks, optimizing athletic performance, diagnosing sources of pain, and monitoring patient recovery trajectories.

\subsection{Interaction Datasets}

Interaction-based motion can be broadly categorized into two forms: (i) interaction between a human body and its surrounding environment or specific objects, and (ii) interaction between two individuals, which can influence the flow of conversation and reflect synchronization, emotional exchange, and interpersonal relationships.

For human–object interaction, GRAB~\cite{taheri2020grab} captures detailed contact information, hand pose, and full-body motion as humans manipulate complex 3D objects, enabling modeling of task-dependent coordination. For human–human interaction, InterHuman~\cite{liang2024intergen} focuses on the dynamics between two people and includes language annotations for text-conditioned motion generation. In the context of conversational body motion, Talking With Hands~\cite{lee2019talking} provides paired speech and upper-body gesture data, emphasizing co-speech gestures and natural hand movements during interaction. Similarly, Body2Hands~\cite{ng2021body2hands} focuses on fine-grained hand and upper-body behaviors, highlighting the importance of hand dynamics and gesture realism for generating expressive conversational motions. These datasets collectively demonstrate the role of body gestures as a complementary communication channel alongside speech, making them valuable resources for studying socially-aware conversational motion generation.

Face motion in this context poses additional challenges due to conversational cues and affective responses. ViCo~\cite{zhou2022responsive} provides an early benchmark for listener facial expressions but is limited in scale. Learning2Listen (L2L)\cite{ng2022learning, ng2023can} expands this direction with large-scale interview-style data incorporating audio, video, and text, though listener behaviors remain constrained in terms of listener demographics and emotion spectrum. Audio2Photoreal~\cite{ng2024audio2photoreal} further extends listener reactions to include hand and upper-body motion. RealTalk~\cite{geng2023affective} advances interaction modeling by offering more diverse conversational topics and emotions and, unlike L2L, features face-to-face dialogues where gaze becomes critical for capturing natural social dynamics.

RECOLA~\cite{ringeval2013introducing} and UDIVA~\cite{palmero2021context} are relatively small datasets that capture different activities between two persons, including conversations on pre-determined topics. While the previously mentioned datasets focus more on conversational topics and listener facial expressions, MMF2F~\cite{lin-etal-2025-predicting} and CCDb+~\cite{deng2025ccdb+} provide backchannel annotations, enabling the learning of more natural listener reactions through non-verbal responses.

\section{Evaluation Metrics}
\label{sec:metrics}

\renewcommand{\arraystretch}{1.2}
\begin{table*}[!htbp]
\centering
\begin{threeparttable}

\caption{Evaluation Metrics for Motion Generation}
\label{tab:metrics_list}

\begin{tabular}{p{2.5cm}p{3cm}p{7cm}>{\centering\arraybackslash}p{3cm}}
\toprule
\textbf{Category} & \textbf{Metric} & \textbf{Purpose} & \textbf{Works} \\
\midrule

\multirow{15}{*}{\parbox{2.5cm}{\centering \textbf{Similarity to\\Ground Truth}}} 
& MPJPE & Mean Per-Joint Position Error; measures Euclidean distance between predicted and reference joints & \cite{zhu2023motionbert,10376684,zhang2024incorporating,zhao2023poseformerv2,pavllo2018quaternet,cheng2021motion} \\

& ADE & Average Displacement Error; quantifies trajectory deviation over time & \cite{10376684,barquero2023belfusion,mohamed2020social,tessler2023calm}\\

& MMADE & Multi-Modal Average displacement error & \cite{10376684}\\

& MMFDE & Multi-Modal Final displacement error & \cite{10376684}\\

& FDE & Final Displacement Error; assesses end-state position accuracy & \cite{lucas2022posegpt,10376684,barquero2023belfusion,mohamed2020social,tessler2023calm}\\

& SSIM & Structural Similarity Index Measure; evaluates perceptual image similarity & \cite{lin2024cyberhost,lee2018stochastic,liu2019liquid,liu2021liquid}\\

& PSNR & Peak Signal-to-Noise Ratio; measures reconstruction quality & \cite{lin2024cyberhost,lee2018stochastic}\\

& L1 Distance & Mean Absolute Distance; computes absolute coordinate differences & \cite{ao2023gesturediffuclip}\\

& P-FD & Paired Fréchet Distance; measures listener-speaker dynamics via distribution distances & \cite{ng2022learning,ng2023can}\\

& PCC & Pearson Correlation Coefficient; quantifies global synchrony & \cite{ng2022learning}\\

\midrule

\multirow{4}{*}{\parbox{2.5cm}{\centering\textbf{Distribution Matching}}}
& FID & Fréchet Inception Distance; compares feature distributions & \cite{lucas2022posegpt,guo2024momask,jiang2023motiongpt,tevet2022human,zhang2023generating,zhang2024motiondiffuse,lu2023humantomato,kalakonda2025morag,zhang2025motionanything,barquero2024seamless,wang2023t2m,ng2022learning,ng2023can} \\

& KID & Kernel Inception Distance; MMD-based comparison & \cite{raab2023modi} \\

& IS & Inception score; evaluates generation quality & \cite{tevet2022motionclip,huang2024humannorm} \\

\midrule

\multirow{4}{*}{\parbox{2.5cm}{\centering\textbf{Diversity and Variability}}}
& Variance & Motion variance; quantifies kinematic variation & \cite{wang2024motiongpt,zhou2024emdm,wang2023t2m,liang2024intergen,zhang2025motionanything,barquero2024seamless} \\

& Diversity & Distinct motion patterns across samples & \cite{wang2024motiongpt,wang2023t2m,lu2023humantomato,zhang2025motionanything} \\

& APD & Average pairwise distance in embedding space & \cite{10376684,barquero2023belfusion,hassan2021stochastic} \\

\midrule

\multirow{8}{*}{\parbox{2.5cm}{\centering\textbf{Semantic Alignment}}}
& R-precision & Retrieval precision; text-motion correspondence & \cite{guo2024momask,wang2024motiongpt,tevet2022human,zhang2023generating,zhang2024motiondiffuse,barquero2024seamless,kalakonda2025morag,zhang2025motionanything,lu2023humantomato} \\

& MM-Dist & Multi-modal matching distance & \cite{guo2024momask,tevet2022human,zhang2023generating,barquero2024seamless,lu2023humantomato} \\

& BLEU & N-gram precision for descriptions & \cite{wang2024motiongpt,guo2022tm2t,chen2024motionllm} \\

& Text Similarity & Embedding-based alignment & \cite{yang2025unimumo} \\

& CLIP Score & Cross-modal similarity & \cite{lin2023being} \\

& Text Retrieval Accuracy & Correct pair identification & \cite{zhou2023ude} \\

\midrule

\multirow{5}{*}{\parbox{2.5cm}{\centering\textbf{Physical Realism}}}
& Contact Accuracy & Ground-contact precision & \cite{yuan2023physdiff,liang2024intergen,taheri2023grip} \\

& Penetration Metrics & Self-intersection avoidance & \cite{liang2024intergen,jiang2024scaling} \\

& Physics Consistency & Newtonian mechanics adherence & \cite{yuan2023physdiff,yu2022physformer} \\

& Foot Skating & Detects unnatural sliding & \cite{xie2023omnicontrol} \\

& Grasp Quality & Hand-object interaction realism & \cite{taheri2023grip,taheri2021goal,GRAB:2020} \\

\midrule

\multirow{5}{*}{\parbox{2.5cm}{\centering\textbf{Temporal Quality}}}
& Peak Jerk & Measures acceleration discontinuities & \cite{barquero2024seamless} \\

& Area Under Jerk & Cumulative motion irregularities & \cite{barquero2024seamless} \\

& Transition Quality & Pose trajectory smoothness & \cite{Lee2023MultiAct} \\

& Smoothness Metrics & Temporal derivative analysis & \cite{jiang2022avatarposer} \\

& Temporal Smoothness & Autocorrelation-based coherence & \cite{taheri2021goal,GRAB:2020} \\

\midrule

\multirow{5}{*}{\parbox{2.5cm}{\centering\textbf{Task-Specific Performance}}}
& Beat Alignment & Dance–music synchronization & \cite{dabral2023mofusion,yang2025unimumo,tseng2023edge,li2024ditto,cheng2025audio2moves} \\

& Instruction Acc. & Command-following precision & \cite{chen2023executing,sheng2024exploring} \\

& Grasp Success & Object manipulation success & \cite{tendulkar2023flex,taheri2021goal,GRAB:2020} \\

& Intention Acc. & Goal-oriented behavior & \cite{diomataris2024wandr} \\

& Control Accuracy & Trajectory tracking precision & \cite{dai2024motionlcm,yao2024moconvq} \\

\midrule

\multirow{3}{*}{\parbox{2.5cm}{\centering\textbf{Subjective Evaluation}}}
& User Studies & Human ratings for naturalness & \cite{wang2024motiongpt,wang2023t2m,barquero2024seamless,lin2024cyberhost,li2024ditto} \\

& MOS & Mean Opinion Score & \cite{yang2023diffusestylegesture} \\

\midrule
\multirow{3}{*}{\parbox{2.5cm}{\centering\textbf{Efficiency and Performance}}}
& Runtime & Processing time & \cite{zhou2024emdm} \\

& Inference Speed & Frames per second & \cite{dai2024motionlcm} \\

& Scalability & Performance vs sequence length & \cite{raab2023modi} \\
\bottomrule
\end{tabular}

\end{threeparttable}
\end{table*}
\renewcommand{\arraystretch}{1.0}
Evaluating generated body and face motions requires a combination of quantitative and qualitative metrics to assess both technical performance and perceptual quality. Quantitative metrics typically examine fidelity to ground truth, motion realism, diversity and cross-modal consistency. Complementing these, qualitative evaluation relies on human judgment, often gathered through user studies and Mean Opinion Score (MOS) ratings, to capture subjective perceptions of naturalness, expressiveness and appropriateness. Since body and facial motion generation encompass multiple tasks with differing objectives, no single evaluation protocol is sufficient for all settings. Therefore, rather than organizing metrics strictly by task, we categorize them according to the aspects of motion quality they measure, such as realism, reconstruction fidelity, diversity, temporal consistency and cross-modal alignment. This organization highlights common evaluation principles shared across different motion generation tasks while also discussing task-specific metrics where appropriate. In this survey, we primarily include evaluation metrics that are widely adopted in existing literature and commonly used for benchmarking and comparative analysis across datasets and models. This focus allows us to establish a consistent evaluation framework and provide readers with the foundational metrics that currently define progress in the field. Metrics that are highly task-specific, rarely used, or limited to isolated works are discussed only when they introduce distinctive evaluation perspectives relevant to particular applications.

\subsection{Similarity to Ground Truth}

Similarity-to-ground-truth metrics evaluate how closely generated outputs match their corresponding references, assessing the fidelity and realism of generated motion, trajectories, and reconstructed frames.

For motion evaluation, Mean Per Joint Position Error (MPJPE) computes the average Euclidean distance between predicted and ground-truth 3D joint coordinates, directly measuring pose accuracy~\cite{zhu2023motionbert, 10376684, zhang2024incorporating, zhao2023poseformerv2, pavllo2018quaternet, cheng2021motion}. For trajectory prediction, Average Displacement Error (ADE) measures the mean Euclidean distance between predicted and ground-truth trajectories across all time steps, while Final Displacement Error (FDE) evaluates this error at the final time step, reflecting long-term prediction accuracy~\cite{lucas2022posegpt, 10376684, barquero2023belfusion, mohamed2020social, tessler2023calm}. Since multiple future motions may be plausible, Multi-Modal ADE (MMADE) and Multi-Modal FDE (MMFDE) compute the minimum ADE/FDE over multiple generated samples, capturing the best matching outcome in stochastic generation settings~\cite{10376684}.

For image reconstruction, L1 loss computes the Mean Absolute Error (MAE) between predicted and ground-truth pixels and is widely used due to its robustness to outliers~\cite{ao2023gesturediffuclip}. To assess perceptual quality, Structural Similarity Index Measure (SSIM) evaluates similarities in luminance, contrast, and structure between images~\cite{lin2024cyberhost, lee2018stochastic, liu2019liquid, liu2021liquid}, whereas Peak Signal-to-Noise Ratio (PSNR) quantifies pixel-level fidelity by measuring the ratio between signal and reconstruction noise, with higher values indicating better quality~\cite{lin2024cyberhost, lee2018stochastic}.

Interaction-centric motion generation additionally requires metrics that capture interpersonal coordination. Paired Fréchet Distance (P-FD) measures the alignment between generated and real listener–speaker behaviors~\cite{ng2022learning, ng2023can}, while the Pearson Correlation Coefficient (PCC) quantifies global synchrony by evaluating the temporal correlation between speaker and listener features, with higher values indicating stronger responsiveness and coordination~\cite{ng2022learning}.


\subsection{Distribution Matching}

Unlike similarity-based metrics that evaluate one-to-one correspondence with ground truth, distribution-based metrics assess whether the overall distribution of generated samples matches that of real data. By capturing global statistical properties, they provide insight into the realism, quality, and diversity of generated motions.

Fréchet Inception Distance (FID) is one of the most widely used metrics for evaluating distribution alignment. It measures the distance between the multivariate Gaussian distributions of real and generated feature representations, with lower values indicating better agreement in terms of both quality and diversity~\cite{lucas2022posegpt,guo2024momask, jiang2023motiongpt, tevet2022human, zhang2023generating, zhang2024motiondiffuse, lu2023humantomato, kalakonda2025morag, zhang2025motionanything, barquero2024seamless, wang2023t2m, ng2022learning, ng2023can}. Since FID assumes Gaussian feature distributions, Kernel Inception Distance (KID) was introduced as an alternative based on polynomial kernel Maximum Mean Discrepancy (MMD), avoiding this assumption and often providing more robust estimates for smaller sample sizes~\cite{raab2023modi}. Similar to FID, lower KID values indicate closer alignment between generated and real data.

Inception Score (IS) evaluates the quality and diversity of generated samples by analyzing the conditional label distributions predicted for them~\cite{tevet2022motionclip, huang2024humannorm}. Higher IS values indicate that individual samples are confidently recognizable while the generated set covers a diverse range of classes. However, because IS considers only generated samples and does not compare them against real data, it cannot directly measure distributional similarity.

Together, FID, KID, and IS provide complementary perspectives on generative performance by assessing distributional alignment, sample quality, and diversity beyond individual ground-truth correspondence.

\subsection{Diversity and Variability}
Beyond quality and realism, generative models should also capture the inherent variability of human motion by producing multiple plausible outputs rather than repeating a limited set of patterns. Diversity metrics evaluate this ability, which is particularly important in motion generation where the same condition can lead to several valid responses.

Diversity measures the variability across generated sequences and reflects how well a model captures the natural richness of human movement~\cite{wang2024motiongpt, zhou2024emdm, wang2023t2m, liang2024intergen, zhang2025motionanything, barquero2024seamless}. Low diversity often indicates mode collapse, where the model generates overly similar motions. Multi-modality diversity further assesses whether a model can generate multiple distinct yet plausible motions for the same input condition, a key property of stochastic generative models~\cite{wang2024motiongpt, wang2023t2m, lu2023humantomato, zhang2025motionanything}.

Complementing these measures, Average Pairwise Distance (APD) quantifies diversity by computing the average distance between embeddings of generated motions. By evaluating the spread of samples in the embedding space, APD provides insight into the coverage of the motion manifold and can capture both intra-class and inter-class variability~\cite{10376684, barquero2023belfusion, hassan2021stochastic}. Higher APD values indicate broader exploration of the motion space.


\subsection{Semantic Alignment}

While diversity metrics assess the range of generated motions, semantic alignment metrics evaluate whether those motions are consistent with the intended context or textual description. These metrics are particularly important in text-to-motion generation, where semantic correctness is as critical as realism and diversity.

R-precision measures retrieval performance by determining whether the correct motion (or text) appears among the top retrieved candidates given its paired text (or motion) query, reflecting the semantic relevance of generated motions~\cite{guo2024momask, wang2024motiongpt, tevet2022human, zhang2023generating, zhang2024motiondiffuse, barquero2024seamless, kalakonda2025morag, zhang2025motionanything, lu2023humantomato}. MM-Dist (Multi-Modal Matching Distance) complements this by computing the distance between motion and text embeddings in a shared representation space, with lower values indicating stronger cross-modal alignment~\cite{guo2024momask, tevet2022human, zhang2023generating, barquero2024seamless, lu2023humantomato}.

Traditional NLP metrics have also been adapted for motion-language tasks. BLEU evaluates n-gram overlap between generated and reference descriptions and is commonly used when generating text from motion or vice versa~\cite{wang2024motiongpt, guo2022tm2t, chen2024motionllm}. Although limited to surface-level matching, it provides a simple and reproducible baseline.

More recently, embedding-based approaches have enabled richer semantic evaluation. Text similarity metrics leverage sentence- or paragraph-level embeddings from large language models to assess semantic coherence beyond lexical overlap~\cite{yang2025unimumo}. Similarly, CLIP Score uses joint vision-language representations learned by CLIP to quantify the alignment between motions and text prompts through visual or learned motion embeddings~\cite{radford2021learning, lin2023being}. Finally, Text Retrieval Accuracy evaluates whether the correct textual description can be retrieved from a generated motion, or vice versa, providing a bidirectional measure of semantic grounding~\cite{zhou2023ude}. Higher retrieval accuracy indicates stronger text–motion correspondence and interpretability.

\subsection{Physical Realism}

While semantic alignment evaluates whether generated motions convey the intended meaning, physical realism assesses whether those motions are biomechanically plausible and consistent with the laws governing human movement. Physically realistic motions should exhibit natural environmental interactions, avoid impossible body configurations, and maintain stable dynamics.

Contact Accuracy measures the correctness of predicted contact points, particularly foot-ground interactions, ensuring that characters make natural contact with the floor without hovering, sliding, or penetration~\cite{yuan2023physdiff, liang2024intergen, taheri2023grip}. To maintain spatial consistency, Penetration Metrics quantify self-intersections and body–object collisions, preventing physically implausible overlaps that can disrupt realism, especially in virtual reality and robotics applications~\cite{liang2024intergen, jiang2024scaling}.

Physical plausibility also depends on realistic motion dynamics. Physics Consistency evaluates whether generated motions adhere to fundamental physical principles, such as gravitational effects and momentum conservation~\cite{yuan2023physdiff, yu2022physformer}. Complementing this, Foot Skating metrics detect artifacts where the feet unnaturally slide across the ground during locomotion or standing, a common indicator of unstable motion generation~\cite{xie2023omnicontrol}.

For human–object interaction tasks, Grasp Quality assesses whether hand poses and contact patterns are physically feasible and geometrically compatible with the manipulated object~\cite{taheri2023grip, taheri2021goal, GRAB:2020}. By evaluating both contact realism and grasp plausibility, it provides a fine-grained measure of physical consistency during object manipulation.

\subsection{Temporal Quality}

While physical realism evaluates the plausibility of individual poses and interactions, temporal quality metrics assess the coherence of motion dynamics over time. They measure whether generated sequences evolve smoothly, maintaining consistent transitions in pose, velocity, and acceleration without abrupt discontinuities.

Peak Jerk and Area Under Jerk are commonly used to quantify motion smoothness. Peak Jerk measures the maximum rate of change of acceleration, with higher values indicating sudden, unnatural movements or visual artifacts such as twitching~\cite{barquero2024seamless}. In contrast, Area Under Jerk integrates jerk over the entire sequence, capturing cumulative temporal irregularities and providing a global assessment of smoothness~\cite{barquero2024seamless}.

Complementing these measures, Transition Quality evaluates the fluidity of pose changes, particularly when switching between actions, ensuring that trajectories remain natural and free from implausible shifts~\cite{Lee2023MultiAct}. Smoothness Metrics further analyze temporal derivatives, such as velocity and acceleration, to identify abrupt variations or synthesis noise introduced during generation~\cite{jiang2022avatarposer}. Together, they assess both local transitions and overall motion continuity.

Finally, Temporal Smoothness examines the consistency of motion patterns using techniques such as autocorrelation and frequency-domain analysis~\cite{taheri2021goal, GRAB:2020}. By evaluating the stability of rhythmic dynamics, it is especially useful for periodic activities such as walking or dancing, providing an additional measure of the natural flow of generated motions.

\subsection{Task-Specific Performance}

Building upon the previously discussed metrics, task-specific metrics evaluate whether generated motions successfully fulfill particular functional objectives. Unlike general measures of realism or plausibility, these metrics assess performance in specialized settings such as dance synthesis, instruction-guided generation, and object manipulation.

Beat Alignment measures the synchronization between generated dance motions and accompanying music, reflecting a model’s ability to respond to temporal audio cues and maintain rhythmic consistency~\cite{dabral2023mofusion, yang2025unimumo, tseng2023edge, li2024ditto, cheng2025audio2moves}. In contrast to Semantic Alignment, which evaluates whether the generated motion broadly matches the meaning or intent of the conditioning input, Instruction Accuracy specifically measures how faithfully the generated motion follows explicit textual or symbolic commands, such as “raise left arm” or “turn right”. Thus, Instruction Accuracy reflects the system’s capability to obey structured and controllable directives rather than general semantic consistency~\cite{chen2023executing, sheng2024exploring}. We categorize Instruction Accuracy under task-specific performance because such evaluations are primarily relevant to controllable or instruction-guided generation settings and are not universally applicable across all motion generation tasks. Together, these metrics assess a model’s responsiveness to external conditions and its ability to generate controllable, context-appropriate motions.

For interaction-oriented tasks, Grasp Success evaluates whether a generated motion successfully reaches and manipulates the target object~\cite{tendulkar2023flex, taheri2021goal, GRAB:2020}. This metric is particularly important in hand-object interaction scenarios, where both execution accuracy and physical feasibility determine task completion.

\subsection{Subjective Evaluation}

Following objective and task-specific evaluations, subjective metrics provide a complementary perspective by assessing how generated motions are perceived by humans. They capture qualitative aspects such as naturalness, expressiveness, appropriateness, and overall preference that are difficult to quantify using objective measures alone.

The most common approach is through User Studies, where participants rate or compare generated motions using criteria such as realism, expressiveness, appropriateness to the input, and overall quality~\cite{wang2024motiongpt, wang2023t2m, barquero2024seamless, lin2024cyberhost, li2024ditto}. These evaluations are typically conducted using pairwise comparisons, ranking tasks, or Likert scales, providing valuable insights into perceptual quality in interactive and real-world settings. Mean Opinion Score (MOS) further summarizes these judgments by averaging participant ratings into a single interpretable measure of perceived quality~\cite{yang2023diffusestylegesture}. Originally developed for speech assessment, MOS has become increasingly common in motion synthesis studies.

The importance of human-centered evaluation has been underscored by the GENEA Challenge series~\cite{yoon2022genea, kucherenko2020genea, kucherenko2023genea}, including the 2020, 2022, and 2023 editions, which established standardized benchmarking protocols for co-speech gesture and conversational motion generation. These challenges highlighted the limitations of relying solely on objective metrics and advocated large-scale perceptual studies to evaluate factors such as naturalness, timing, human-likeness, and appropriateness to speech. As a result, they have significantly shaped evaluation practices in conversational and socially aware motion synthesis research.

\begin{figure*}[!ht]
\centering
    \includegraphics[width=\textwidth, height=320pt]{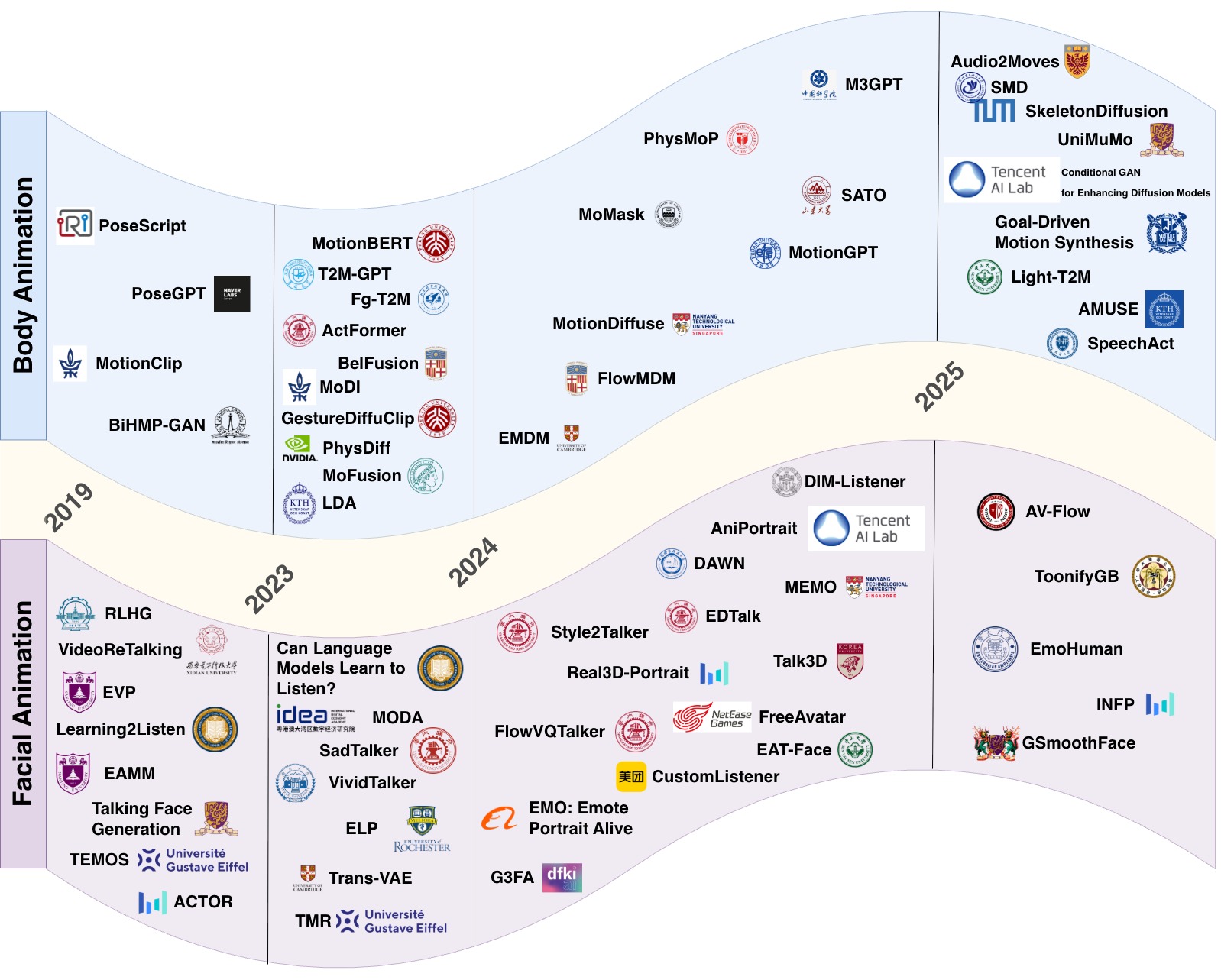}
    \caption{Roadmap of Motion Generation Techniques}
    \label{fig:roadmap}
\end{figure*}

\subsection{Efficiency and Performance}

While subjective evaluations provide insights into human perceptions of motion quality, they do not account for the computational feasibility of generating such motions. Efficiency and performance metrics evaluate the computational aspects of motion generation models, focusing on their practicality for deployment in real-time or large-scale applications. These metrics are particularly important for interactive systems, edge devices or scenarios requiring low-latency responses.

Runtime measures the overall computational latency involved in processing a single input or a batch of inputs, capturing the end-to-end time required for motion generation~\cite{zhou2024emdm}. It serves as a key indicator of the model’s practical usability in time-sensitive applications such as virtual avatars or robotics. Complementing this, Inference Speed, typically expressed in frames per second (FPS), evaluates how efficiently the model generates motion during inference~\cite{dai2024motionlcm}. A higher FPS signifies that the system can produce motion at or above real-time rates, ensuring responsiveness and smooth performance in interactive or streaming scenarios.

In addition to runtime and inference speed, Scalability evaluates how the model’s performance, both in terms of speed and memory consumption, scales with increasing input size, such as longer sequence lengths or higher dimensional inputs~\cite{raab2023modi}. This metric is important for understanding the model’s efficiency under varying workload conditions and ensuring robustness when deployed in dynamic or large-scale environments.

\section{Approaches}

Motion generation methods primarily focus on either facial motion or body motion. Despite this distinction, they share common modeling principles and can generally be categorized into two groups: \textbf{(1) Conventional Generative Models} and \textbf{(2) Diffusion-based Models}. Conventional generative models directly learn mappings from input modalities to motion representations using architectures such as GANs, VAEs, autoregressive models and transformer-based generators. In contrast, diffusion-based models synthesize motion through an iterative denoising process, progressively refining noisy samples into realistic motion sequences. Figure~\ref{fig:pipeline} illustrates a general motion generation pipeline, which can accommodate a wide range of input modalities. For clarity and ease of comparison, we further organize existing approaches according to their target motion domain, namely face animation and body animation, as summarized in Table~\ref{tab:face_list} and Table~\ref{tab:body_list}, respectively. Since 2023, motion generation research has experienced rapid growth, particularly driven by the emergence and success of diffusion-based approaches, as illustrated in Figure~\ref{fig:roadmap}.

\subsection{Face Animation}
\begin{table*}[!t]
\centering
\begin{threeparttable}

\caption{Face Animation Methods}
\label{tab:face_list}

\begin{tabular}{p{2cm}p{3cm}c>{\centering\arraybackslash}p{2cm}p{8.5cm}}
\toprule
\textbf{Category} & \textbf{Approach} & \textbf{Year} & \textbf{Input} & \textbf{Main Contribution}\\
\midrule

\multirow{10}{*}{\parbox{2cm}{\centering\textbf{Diffusion-based Models}}}

& AV-Flow~\cite{chatziagapi2025av} & 2025 & \texticon & Generate Audio-Visual content \\

& AniPortrait~\cite{wei2024aniportrait} & 2024 & \audioicon + \imageicon & High-quality animation driven by audio and a reference portrait image \\

& DAWN~\cite{cheng2024dawn} & 2024 & \audioicon + \imageicon & Audio-driven facial motion, head pose and blink generation \\

& EDTalk~\cite{tan2024edtalk} & 2024 & \audioicon + \imageicon & Distinct representation of lip, head pose and emotional expression \\

& EMO: Emote Portrait Alive~\cite{tian2024emo} & 2024 & \audioicon + \imageicon & Expressive audio-driven portrait-video generation \\

& MEMO~\cite{zheng2024memo} & 2024 & \audioicon + \imageicon & Identity-consistent and expressive talking videos \\

& Real3D-Portrait~\cite{ye2024real3d} & 2024 & \audioicon + \videoicon + \imageicon & One-shot video/audio-driven 3D talking face generation \\

& EAT-Face~\cite{wang2024eat} & 2024 & \audioicon + \imageicon & Emotion-controllable audio-driven talking face generation \\

\midrule

\multirow{33}{*}{\parbox{2cm}{\centering\textbf{Conventional Generative Models}}}

& ToonifyGB~\cite{ju2025toonifygb} & 2025 & \videoicon & Stylized avatars with arbitrary expressions using Gaussian blendshapes \\

& GSmoothFace~\cite{zhang2025gsmoothface} & 2025 & \audioicon + \videoicon & Improves 3D face reconstruction/animation by enforcing geometric and temporal smoothness to produce more stable and realistic facial motion \\

& EmoHuman~\cite{dai2025emohuman} & 2025 & \audioicon + \imageicon & Controllable expressive talking head generation \\

& G3FA~\cite{javanmardi2024g3fa} & 2024 & \videoicon + \imageicon & Geometry-guided facial animation with 3D cues \\

& DIM-Listener~\cite{tran2024dim} & 2024 & \audioicon + \videoicon & Dynamic Interaction Modeling to generate both speaker and listener motion \\

& Style2Talker~\cite{tan2024style2talker} & 2024 & \audioicon + \imageicon + \texticon & Audio-driven face generation with text-controlled emotion \\

& CustomListener~\cite{liu2024customlistener} & 2024 & \texticon + \audioicon & Text-guided listener generation \\

& Talk3D~\cite{ko2024talk3d} & 2024 & \audioicon + \motionicon & Audio-driven head generation with 3D-aware priors \\

& FlowVQTalker~\cite{tan2024flowvqtalker} & 2024 & \audioicon + \imageicon & Flow + VQ modeling for emotional talking faces \\

& FreeAvatar~\cite{qiu2024freeavatar} & 2024 & \imageicon & High-fidelity 3D avatars from in-the-wild images \\

& Can Language Models Learn to Listen?~\cite{ng2023can} & 2023 & \texticon & Listener facial responses from speech content \\

& MODA~\cite{liu2023moda} & 2023 & \audioicon + \imageicon & Audio-driven talking face with dual attention \\

& SadTalker~\cite{zhang2023sadtalker} & 2023 & \audioicon + \imageicon & Realistic talking videos from audio + image \\

& VividTalk~\cite{sun2023vividtalk} & 2023 & \audioicon + \imageicon & Mesh-to-animation talking head generation \\

& ELP~\cite{song2023emotional} & 2023 & \audioicon + \videoicon & Emotional latent-space listener expressions \\

& Trans-VAE~\cite{song2023react2023} & 2023 & \audioicon + \videoicon & Multiple appropriate facial reactions \\

& TMR~\cite{Petrovich_2023_ICCV} & 2023 & \texticon & Text-motion alignment by introducing contrastive learning into the TEMOS framework for improved text-to-3D human motion retrieval.\\

& VideoReTalking~\cite{cheng2022videoretalking} & 2022 & \audioicon + \videoicon & Photo-realistic talking video with lip-audio synchronization \\

& EAMM~\cite{ji2022eamm} & 2022 & \audioicon + \videoicon & Talking face matching audio and emotional dynamics \\

& Learning2Listen~\cite{ng2022learning} & 2022 & \audioicon + \videoicon & Listener response generation with Motion VQ-VAE \\

& RLHG~\cite{zhou2022responsive} & 2022 & \audioicon + \videoicon & Responsive listener feedback synthesis \\

& TEMOS~\cite{petrovich2022temos} & 2022 & \texticon & Text-driven latent motion framework that aligns language and motion representations to generate human motions from natural language descriptions. \\

& EVP~\cite{ji2021audio} & 2021 & \audioicon + \videoicon & Emotion-controllable talking face \\

& ACTOR~\cite{Petrovich_2021_ICCV} & 2021 & \texticon & A Transformer-based VAE for generating human motions from action-conditioned latent representations without requiring an initial pose. \\

& Talking Face Generation~\cite{zhou2019talking} & 2019 & \audioicon + \videoicon & Speech-driven facial image synthesis \\

\bottomrule
\end{tabular}
\vspace{3pt}
\begin{tablenotes}
\footnotesize
\item \texticon: Text; \audioicon: Audio; \videoicon: Video (Excludes audio); \imageicon: Image; \motionicon: Motion
\end{tablenotes}

\end{threeparttable}
\end{table*}

\subsubsection{Diffusion Model}
Recent works have increasingly focused on diffusion-based approaches for audio and video-driven animation. AniPortrait~\cite{wei2024aniportrait} achieves high-quality portrait animation from audio using only a single reference image. DAWN~\cite{cheng2024dawn} extends this idea by introducing the Pose and Blink generation Network (PBNet), which enhances realism through synchronized face motion, head pose and blinking. EDTalk~\cite{tan2024edtalk} provides greater control by disentangling lip motion, head pose and emotional expression, while EAT-Face~\cite{wang2024eat} further enables fine-grained control through emotion-controllable talking face generation from audio and image inputs.

Other approaches emphasize expression richness and identity preservation. EMO: Emote Portrait Alive~\cite{tian2024emo} and MEMO~\cite{zheng2024memo} generate expressive and identity-consistent portrait videos, maintaining fidelity to the reference image. Extending to 3D synthesis, Real3D-Portrait~\cite{ye2024real3d} supports realistic one-shot 3D talking face generation from a reference image, driven not only by audio but also alternatively by video. Distinct from these audio and image-based methods, AV-Flow~\cite{chatziagapi2025av} leverages text input to jointly generate both face animation and synthesized speech, broadening the scope of multi-modal diffusion frameworks.

\subsubsection{Conventional Generative Models}
Conventional generative models have played a foundational role in face animation by directly learning mappings between input modalities and facial motion representations. Early GAN-based approaches focused on generating realistic and temporally coherent talking faces from speech. Talking Face Generation~\cite{zhou2019talking} represents one of the earliest methods to synthesize face image sequences directly from audio, while VideoReTalking~\cite{cheng2022videoretalking} improves realism by re-synchronizing lip movements with alternative speech while preserving natural facial expressions. EAMM~\cite{ji2022eamm} further incorporates emotional cues from reference videos to produce emotion-aware talking faces. More recent GAN-based methods have emphasized controllability and stylization. Style2Talker~\cite{tan2024style2talker} enables text-guided emotional control and image-based artistic styles for audio-driven talking face generation, whereas G3FA~\cite{javanmardi2024g3fa} leverages 3D geometric priors to animate realistic faces from a single image. ToonifyGB~\cite{ju2025toonifygb} further advances stylized animation by integrating Gaussian blendshapes within a GAN framework to render expressive avatars with rich visual details.

Alongside GAN-based approaches, VAE-based and other neural generative architectures have been widely adopted for controllable facial motion synthesis. EVP~\cite{ji2021audio} explores emotion-controllable talking face generation conditioned on both audio and target video inputs, laying the groundwork for emotion-aware synthesis. Subsequent methods focus on improving realism, expressiveness, and multi-modal control. MODA~\cite{liu2023moda} captures subject-specific speaking styles through dual attention mechanisms, while SadTalker~\cite{zhang2023sadtalker} generates synchronized talking videos from a single reference image and audio. VividTalk~\cite{sun2023vividtalk} employs a dual-branch Motion-VAE to transform mesh-based facial representations into expressive and temporally coherent animations. Building on these foundations, Talk3D~\cite{ko2024talk3d} reconstructs detailed facial geometries using pre-trained 3D-aware priors, FlowVQTalker~\cite{tan2024flowvqtalker} combines normalizing flows and vector quantization for emotion-rich talking face synthesis, FreeAvatar~\cite{qiu2024freeavatar} generates high-fidelity 3D avatars from in-the-wild images, and EmoHuman~\cite{dai2025emohuman} disentangles content, emotion, and intensity to enable fine-grained expressive control.

Conventional generative models have also been extensively explored for listener face animation. Learning2Listen~\cite{ng2022learning} employs a Motion VQ-VAE to generate diverse listener reactions, while RLHG~\cite{zhou2022responsive} synthesizes responsive gestures such as nods and smiles. Can Language Models Learn to Listen?\cite{ng2023can} introduces text-guided listener response generation, and ELP\cite{song2023emotional} and Trans-VAE~\cite{song2023react2023} further enhance emotional expressiveness through latent emotion modeling and blink-aware generation mechanisms.

\subsection{Body Animation}
\begin{table*}[!htbp]
\centering
\begin{threeparttable}

\caption{Body Animation Methods}
\label{tab:body_list}

\begin{tabular}{p{2cm}p{3cm}c>{\centering\arraybackslash}p{2cm}p{8.5cm}}
\toprule
\textbf{Category} & \textbf{Approach} & \textbf{Year} & \textbf{Input} & \textbf{Main Contribution}\\
\midrule

\multirow{17}{*}{\parbox{2cm}{\centering\textbf{Diffusion-based Models}}}

& SMD~\cite{wu2025doodle} & 2025 & \motionicon & Generates 3D body motion based on 2D Sketch information \\

& Goal-Driven Motion Synthesis~\cite{hwang2025goal} & 2025 & \motionicon & Given initial and target poses, generate intermediate motion \\

& Light-T2M~\cite{zeng2025light} & 2025 & \texticon & Lightweight Text-to-Motion generation model \\

& AMUSE~\cite{chhatre2024emotional} & 2025 & \audioicon & Emotional speech-driven body animation \\

& SkeletonDiffusion~\cite{curreli2025nonisotropic} & 2025 & \motionicon & Given past motion, generate plausible and semantically coherent future motion \\

& UniMuMo~\cite{yang2025unimumo} & 2025 & \audioicon + \texticon + \motionicon & Unified multi-modal model capable of taking arbitrary text, music and motion data as input conditions to generate outputs across all three modalities \\

& EMDM~\cite{zhou2024emdm} & 2024 & \texticon & Generate fast and diverse motion from given text input \\

& FlowMDM~\cite{barquero2024seamless} & 2024 & \texticon & Seamlessly generate human motion from text description \\

& MotionDiffuse~\cite{zhang2024motiondiffuse} & 2024 & \texticon & Probabilistic mapping to generate diverse motion from given input text \\

& MoFusion~\cite{dabral2023mofusion} & 2023 & \audioicon + \texticon & Generate motion conditioned on text and audio, matching music rhythm \\

& PhysDiff~\cite{yuan2023physdiff} & 2023 & \texticon + \motionicon & Generate physics-aware motion \\

& GestureDiffuCLIP~\cite{ao2023gesturediffuclip} & 2023 & \audioicon + \texticon & Speech-conditioned gesture generation with styling features \\

& LDA~\cite{10.1145/3592458} & 2023 & \audioicon & Conformers-based framework for probabilistic audio-driven human motion generation with controllable motion style synthesis.\\

\midrule

\multirow{20}{*}{\parbox{2cm}{\centering\textbf{Conventional Generative Models}}}

& Conditional GAN for Enhancing Diffusion Models~\cite{cheng2025conditional} & 2025 & \audioicon & Co-speech gesture generation from audio using conditional GAN + diffusion \\

& SpeechAct~\cite{zhang2025speechact} & 2025 & \audioicon & Generates diverse body and face motion based on speech \\

& Audio2Moves~\cite{cheng2025audio2moves} & 2025 & \audioicon & Generates motion synchronized with audio beats \\

& MotionGPT~\cite{ribeiro2024motiongpt} & 2024 & \texticon & Uses LLM semantic priors for detailed motion generation \\

& MoMask~\cite{guo2024momask} & 2024 & \texticon & High-quality 3D human motion with diversity and control \\

& SATO~\cite{chen2024sato} & 2024 & \texticon & Addresses instability in Text-to-Motion models \\

& M$^{3}$GPT~\cite{luo2024m} & 2024 & \audioicon + \texticon + \motionicon & Discretizes multi-modal signals into unified motion space \\

& PhysMoP~\cite{zhang2024incorporating} & 2024 & \motionicon & Physics-aware motion with long-term dependency modeling \\

& MoDI~\cite{raab2023modi} & 2023 & \motionicon & Unsupervised generation of diverse motion from a given distribution \\

& BelFusion~\cite{barquero2023belfusion} & 2023 & \motionicon & Samples from behavioral latent space disentangled from pose and motion \\

& ActFormer~\cite{xu2023actformer} & 2023 & \texticon & Action-conditioned motion Transformer trained with GAN scheme \\

& Fg-T2M~\cite{wang2023fg} & 2023 & \texticon & Context-aware progressive reasoning for fine-grained motion \\

& T2M-GPT~\cite{zhang2023generating} & 2023 & \texticon & Generates discrete motion representations from text \\

& MotionBERT~\cite{zhu2023motionbert} & 2023 & \videoicon & Multi-task generation of 3D pose, mesh and action labels \\

& MotionCLIP~\cite{tevet2022motionclip} & 2022 & \texticon & Projects motion manifold into CLIP latent space \\

& PoseGPT~\cite{lucas2022posegpt} & 2022 & \texticon + \motionicon & Motion generation conditioned on action and duration \\

& BiHMP-GAN~\cite{kundu2019bihmp} & 2019 & \motionicon & Bi-directional framework to generate motion from a starting sequence \\

& PoseScript~\cite{delmas2022posescript} & 2022 & \texticon & Associates text descriptions with pose codes \\

\bottomrule
\end{tabular}

\vspace{3pt}
\begin{tablenotes}
\footnotesize
\item \texticon: Text; \audioicon: Audio; \videoicon: Video (Excludes audio); \imageicon: Image; \motionicon: Motion
\end{tablenotes}

\end{threeparttable}
\end{table*}

\subsubsection{Diffusion Model}
Early approaches focus on text and audio-driven human motion generation, aiming to produce synchronized and context-aware movements. MoFusion~\cite{dabral2023mofusion} generates motion conditioned on both text and audio, aligning body dynamics with musical rhythm, while PhysDiff~\cite{yuan2023physdiff} integrates physics-based constraints to ensure physically plausible motion synthesis. GestureDiffuCLIP~\cite{ao2023gesturediffuclip} extends this direction towards speech-conditioned gesture generation, incorporating stylistic control for more expressive outputs. Building on these foundations, subsequent research focuses towards faster, more diverse and semantically rich motion synthesis. Models such as MotionDiffuse~\cite{zhang2024motiondiffuse}, FlowMDM~\cite{barquero2024seamless} and EMDM~\cite{zhou2024emdm} leverage diffusion and flow-based probabilistic modeling to generate text-conditioned motions that are both seamless and diverse. More recent works emphasize goal-directed, multi-modal and long-sequence motion generation. Goal-Driven Motion Synthesis~\cite{hwang2025goal} focuses on producing smooth transitions between given start and target poses, while SkeletonDiffusion~\cite{curreli2025nonisotropic} predicts future motions from past skeleton sequences to ensure temporal coherence. Light-T2M~\cite{zeng2025light} introduces an efficient and lightweight text-to-motion framework. AMUSE~\cite{chhatre2024emotional} specializes in emotionally expressive speech-driven body animation and UniMuMo~\cite{yang2025unimumo} unifies multi-modal conditioning to flexibly generate motion from text, music or skeleton inputs.

\subsubsection{Conventional Generative Models}

Conventional generative models have established the foundation of modern body motion generation by directly learning motion distributions through adversarial learning, latent variable modeling, and sequence generation frameworks. Early GAN-based approaches focused on synthesizing realistic and diverse motion sequences. BiHMP-GAN~\cite{kundu2019bihmp} introduced a bi-directional adversarial framework for generating future human motion conditioned on observed motion history, while MoDi~\cite{raab2023modi} employed an unsupervised GAN architecture to model and generate diverse skeleton motions from a learned motion distribution. Extending adversarial learning to interactive settings, ActFormer~\cite{xu2023actformer} combines Transformer architectures, GAN training, and Gaussian process latent priors to synthesize action-conditioned multi-person motions. More recent works have explored hybrid paradigms that combine adversarial learning with diffusion-based representations. For example, BelFusion~\cite{barquero2023belfusion} integrates adversarial training with latent diffusion representations to disentangle behavioral factors from pose and motion, while Conditional GAN Diffusion~\cite{cheng2025conditional} incorporates conditional GANs into diffusion frameworks to generate realistic co-speech gestures with reduced sampling complexity.

Alongside GAN-based approaches, latent representation learning and neural generative architectures have become increasingly influential. ACTOR~\cite{petrovich2021action} introduced a Transformer-based variational framework capable of generating diverse motions conditioned on action labels, demonstrating the effectiveness of latent variable modeling for motion synthesis. Building on this idea, TEMOS~\cite{petrovich2022temos} learns a shared latent space between text and motion through a variational framework, enabling direct text-driven motion generation, while TMR~\cite{Petrovich_2023_ICCV} further improves text-motion alignment through a retrieval-oriented joint representation space. MotionCLIP~\cite{tevet2022motionclip} leverages pre-trained vision-language models to embed human motion into a semantically meaningful latent space, facilitating controllable motion generation and editing. Similarly, PoseGPT~\cite{lucas2022posegpt} formulates motion generation as a sequence modeling task conditioned on action labels, duration, and observed motion, while PoseScript~\cite{delmas2022posescript} establishes correspondences between natural language descriptions and discrete human pose representations.

Subsequent research has focused on improving semantic understanding, controllability, and multi-modal integration. Fg-T2M~\cite{wang2023fg} introduces context-aware progressive reasoning to generate fine-grained motions from textual descriptions, whereas T2M-GPT~\cite{zhang2023generating} employs autoregressive modeling to synthesize discrete motion tokens from text. MotionBERT~\cite{zhu2023motionbert} extends motion modeling to a unified multi-task framework capable of generating 3D poses, meshes, and action labels from video inputs. More recent methods further exploit large-scale language and multi-modal models for motion generation. MoMask~\cite{guo2024momask} generates diverse and controllable 3D human motions from text, while MotionGPT~\cite{ribeiro2024motiongpt} leverages semantic knowledge encoded in large language models to produce detailed and semantically coherent motion sequences. SATO~\cite{chen2024sato} addresses instability issues in text-to-motion generation, and M$^3$GPT~\cite{luo2024m} unifies text, audio, and skeleton motion within a shared discrete representation space to support flexible multi-modal generation. Furthermore, PhysMoP~\cite{zhang2024incorporating} incorporates physics-aware constraints and long-term temporal modeling to improve motion realism, while Audio2Moves~\cite{cheng2025audio2moves} generates body motions synchronized with audio beats while preserving the stylistic characteristics of the input audio.

\section{Conclusion and Future Directions}

\begin{figure*}[!ht]
    \centering
    \includegraphics[width=1.0\linewidth]{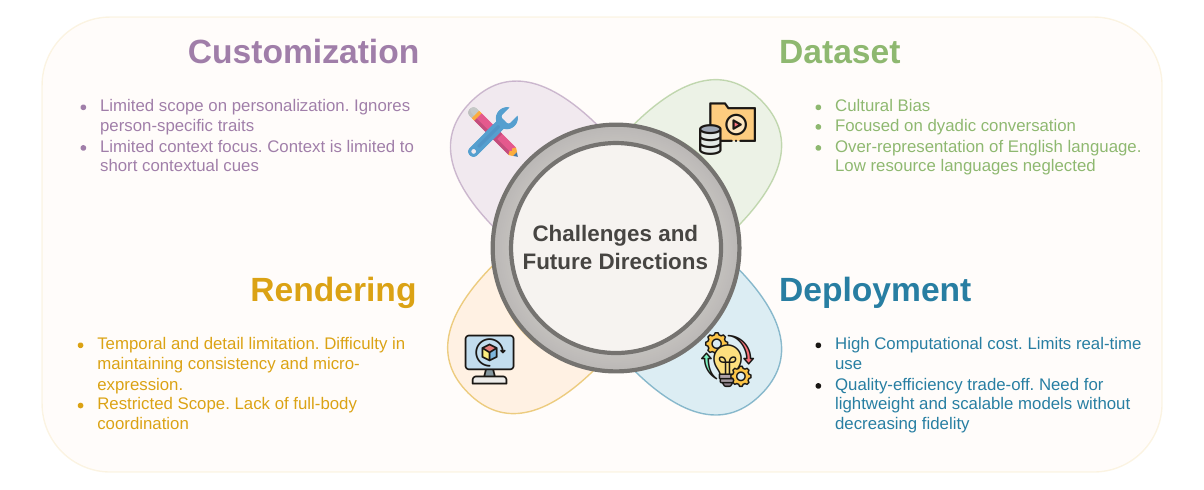}
    \caption{Overview of Key Challenges and Emerging Future Research Directions}
    \label{fig:challenges}
\end{figure*}

In this survey, we present a comprehensive overview of recent advancements in human body and face motion generation. We provide a detailed comparison of existing datasets, highlighting the modalities, task categories and representation methods. We summarize the evaluation metrics commonly employed for both body and face animation. To structure the discussion, we examine body and face motion separately, emphasizing the variety of task formulations and organizing existing methods into three main categories: Diffusion-based, GAN-based, and Neural Network/VAE-based approaches. Building upon these insights, this section discusses the key challenges that remain and outlines promising future research directions in this rapidly evolving field, as illustrated in Figure~\ref{fig:challenges}.

\subsubsection{Dataset} 
While Audio2Photoreal~\cite{ng2024audio2photoreal} introduces a full-body motion dataset to bridge the gap in generating both speaker and listener motion, its limited participant pool, consisting of only four individuals, restricts the diversity and generalizability of human behaviors captured. This constraint restricts modeling of complex human behavior, which inherently varies across individual traits and alternative interaction patterns. Furthermore, most existing datasets remain restricted to dyadic scenarios, leaving the extension to group interactions, where the dynamics of multiple participants play a crucial role, largely unexplored. From a linguistic perspective, most datasets primarily feature English speech and text, with some recent efforts in Mandarin~\cite{liu2025disenttalk}. Nevertheless, many low-resource languages remain under-represented, despite their unique linguistic and cultural characteristics. Bridging this gap would advance the modeling of human motion in ways that better reflect cultural diversity and inclusivity.

\subsubsection{Deployment} 
Current motion generation techniques primarily emphasize diversity, realism and controllability, but these qualities often come at the expense of high computational cost. Such requirements hinder their deployment in real-time or resource-constrained environments, such as mobile devices, AR/VR headsets or interactive social robots. A promising direction for future research is the development of lightweight generation techniques that maintain fidelity while reducing computational overhead. This involves model compression, knowledge distillation and other efficient architectures tailored for low-latency generation. Furthermore, balancing the trade-off between quality and efficiency remains an open challenge. Methods capable of adaptively scaling computation based on task demands may prove especially valuable. Exploring these directions will broaden the practical applicability of motion generation systems and enable their integration into day-to-day interactive technologies.

\subsubsection{Rendering}
Frameworks such as PIRenderer~\cite{ren2021pirenderer} leverage 3DMM coefficients to synthesize realistic face motions from reference images, significantly enhancing the realism and naturalness of face animation. While effective for portrait-level synthesis, these approaches still face challenges in generating high-fidelity, temporally consistent image sequences, particularly when capturing subtle dynamics such as micro-expressions, lighting variations and fine-grained texture details. In addition, real-time streaming and deployment remain challenging, as many state-of-the-art rendering and animation frameworks are computationally heavy and difficult to execute under low-latency constraints required for interactive applications. Moreover, current efforts remain largely restricted to the face and upper body regions. Extending such methods to full-body avatars, capable of modeling complex poses, gestures and body–face coordination, remains an important yet underexplored avenue. Achieving this would require balancing parametric control with photorealistic rendering to create animations that are not only realistic at the frame level but also coherent across sequences. Progress in this direction will allow the development of holistic human motion generation systems that integrate both facial expressiveness and body articulation, enabling more immersive applications in virtual humans, digital avatars and human–computer interaction.

\subsubsection{Avatar customization} 
While recent studies have explored customization through text descriptions~\cite{liu2024customlistener, tan2024flowvqtalker, song2023emotional}, they generally do not account for user-specific personality traits. Existing approaches tend to focus on short, context-dependent descriptions tied to conversations and anticipated reactions, without modeling deeper individual differences. A promising future direction lies in extending these methods to capture and learn personality traits, such as extroversion, agreeableness, or emotional expressiveness, which could enable animation styles that are not only context-aware but also aligned with the unique behavioral characteristics of users. This would contribute to generating more natural, consistent and human-like motion behaviors.




\bibliographystyle{IEEEtran}
\bibliography{references.bib}

\end{document}